\documentclass[10pt,twocolumn,letterpaper]{article}

\usepackage{cvpr}              

\usepackage{graphicx}
\usepackage{amsmath}
\usepackage{amssymb}
\usepackage{booktabs}
\usepackage{kotex}
\usepackage{url}            
\usepackage{amsfonts}       
\usepackage{amssymb}
\usepackage{nicefrac}       
\usepackage{microtype}      
\usepackage{adjustbox}
\usepackage{multirow}
\usepackage[flushleft]{threeparttable}
\usepackage{makecell}
\usepackage{tabulary}
\usepackage{amsmath}
\usepackage{tabularx}
\usepackage{pifont}
\usepackage{kotex}
\usepackage{bbold}
\usepackage{wrapfig}

\usepackage{enumitem}
\usepackage{graphicx}
\usepackage{scalerel}
\usepackage{caption}
\usepackage{tabularx}
\usepackage[linesnumbered,ruled,vlined]{algorithm2e}
\usepackage[pagebackref,breaklinks,colorlinks]{hyperref}

\usepackage[capitalize]{cleveref}
\crefname{section}{Sec.}{Secs.}
\Crefname{section}{Section}{Sections}
\Crefname{table}{Table}{Tables}
\crefname{table}{Tab.}{Tabs.}

\newcommand{\cmark}{\text{\ding{51}}}

\def\cvprPaperID{573} 
\def\confName{CVPR}
\def\confYear{2022}

\begin{document}

\title{Weakly Supervised Semantic Segmentation using Out-of-Distribution Data}

\author{Jungbeom Lee$^1$ ~ Seong Joon Oh$^{2, 3}$ ~ Sangdoo Yun$^2$ ~ Junsuk Choe$^4$ ~ Eunji Kim$^1$ ~  Sungroh Yoon$^{1, 5,}$\thanks{Correspondence to: Sungroh Yoon (sryoon@snu.ac.kr).}\\
\small
$^1$Department of Electrical and Computer Engineering, Seoul National University\\
\small
$^2$NAVER AI Lab ~~~~~~~~~~
$^3$University of T\"ubingen\\
\small
$^4$Department of Computer Science and Engineering, Sogang University\\
\small
$^5$Interdisciplinary Program in AI, AIIS, ASRI, INMC, ISRC and NSI, Seoul National University\\
}
\maketitle

\begin{abstract}
Weakly supervised semantic segmentation (WSSS) methods are often built on pixel-level localization maps obtained from a classifier. However, training on class labels only, classifiers suffer from the spurious correlation between foreground and background cues (e.g. train and rail), fundamentally bounding the performance of WSSS. 
There have been previous endeavors to address this issue with additional supervision.
We propose a novel source of information to distinguish foreground from the background: \textbf{Out-of-Distribution (OoD) data}, or images devoid of foreground object classes. 
In particular, we utilize the \textbf{hard OoDs} that the classifier is likely to make false-positive predictions.
These samples typically carry key visual features on the background (e.g. rail) that the classifiers often confuse as foreground (e.g. train), so these cues let classifiers correctly suppress spurious background cues.
Acquiring such hard OoDs does not require an extensive amount of annotation efforts; it only incurs a few additional image-level labeling costs on top of the original efforts to collect class labels.
We propose a method, \textbf{W-OoD}, for utilizing the hard OoDs. 
W-OoD achieves state-of-the-art performance on Pascal VOC 2012.
The code is available at: \url{https://github.com/naver-ai/w-ood}.
\end{abstract}

\vspace{-1em}
\section{Introduction}
\label{sec:intro}
Pixel-wise labeling is labor-intensive \cite{cordts2016cityscapes}. Lots of research have been dedicated to supervising a semantic segmentation model with weaker forms of supervision than pixel-wise labelings, such as scribbles~\cite{tang2018normalized}, points~\cite{bearman2016s, kim2021beyond}, boxes~\cite{khoreva2017simple,song2019box, lee2021bbam}, and class labels~\cite{wang2020self, lee2021anti, lee2021reducing, lee2018robust}. 
We tackle the final category in this paper: weakly supervised semantic segmentation (WSSS) with class labels. 

WSSS methods utilizing class labels often follow a two-stage process. First, they generate pixel-level pseudo-target from a classifier using CAM variants~\cite{zhou2016learning,selvaraju2017grad}. Then, they train the main segmentation network using the pseudo-target generated in the first stage. 
Built on image-level labels only, the pseudo-target is known to suffer from the confusion between foreground and background cues. For example, given a database of duck images where ducks are typically waterborne, a classifier erroneously assigns higher scores on patches containing water than those containing ducks' feet \cite{choe2020evaluating,li2019guided, zhang2020causal, lee2021reducing, lee2021railroad, kolesnikov2016improving}. The same goes for foreground-background pairs like woodpecker-tree, snowmobile-snow, and train-rail. This is a fundamental problem that cannot be solved solely with the class labels; additional information is needed to learn to fully distinguish the foreground and background cues \cite{choe2020evaluating, li2019guided, lee2021railroad}.

\begin{figure}[t]
\centering
\includegraphics[width=\linewidth]{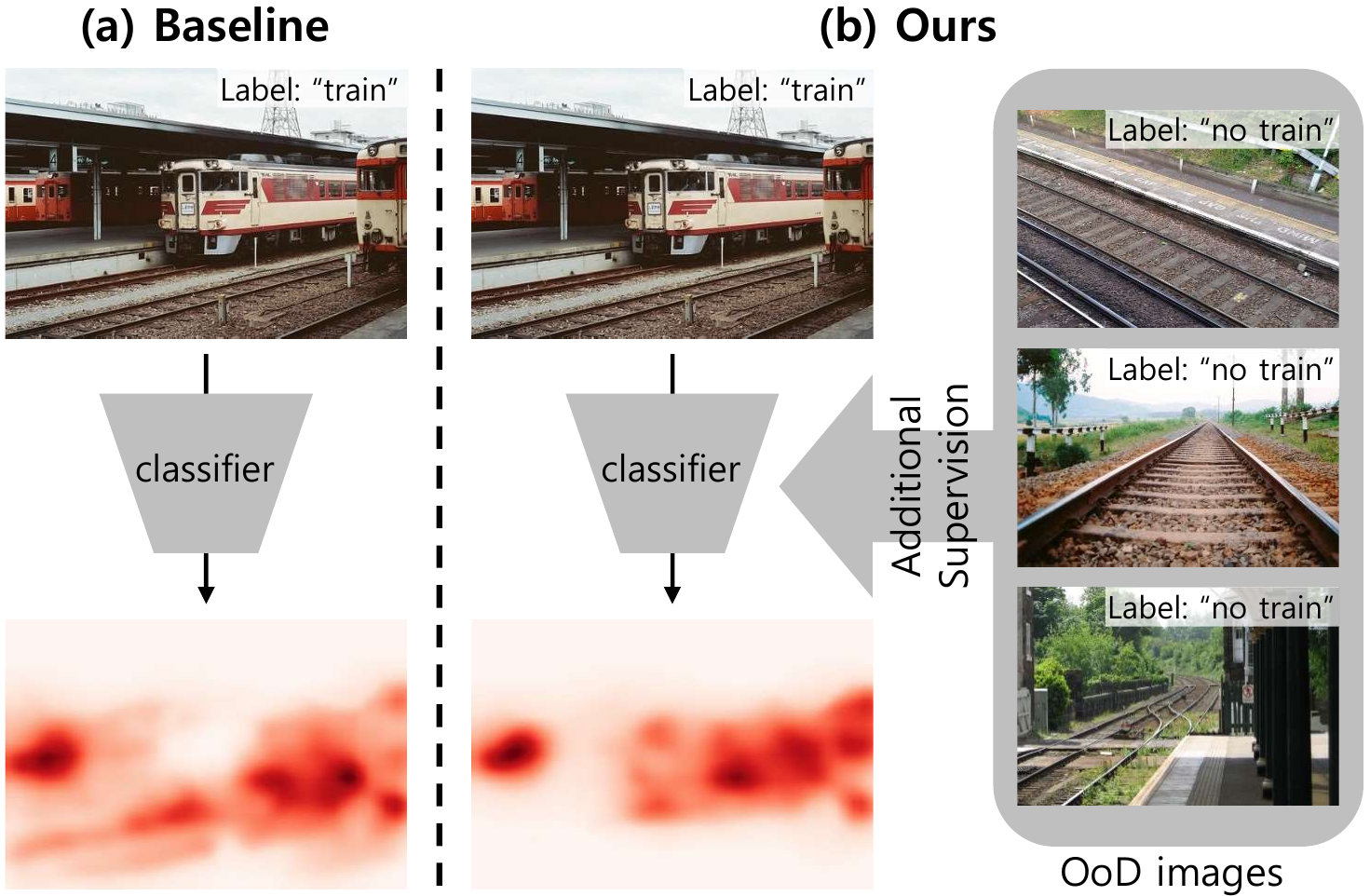}
\vspace{-2em}
\caption{\label{fig1} (a) Classifiers often confuse background cues to be a foreground concept due to spurious correlations (\eg ``rail'' for ``train''). (b) Our W-OoD employs hard OoD images as negative samples (\eg ``rail'' is not ``train'') to resolve the confusion.}
\vspace{-1.3em}
\end{figure}

Researchers have thus sought various sources of additional guidance to separate the foreground and background cues, each with different pros and cons and different labeling-cost footprints. Image saliency~\cite{li2014secrets, liu2010learning} is one of the most widely used ones \cite{lee2019ficklenet, lee2021railroad, yao2021nonsalient, sun2020mining, liu2020leveraging, joon2017exploiting}, for it naturally provides the prominent foreground object in the image in a class-agnostic fashion. However, saliency is not very effective for non-salient foreground objects (\eg low-contrast objects or small objects). Low-level visual features like superpixels~\cite{kwak2017weakly, wang2018weakly}, edges~\cite{ke2021universal}, object proposals~\cite{lee2021bbam, song2019box, liu2020leveraging}, and optical flows~\cite{hong2017weakly, lee2019frame} have also been considered. 
Though cost-effective, they tend to generate inaccurate object boundaries because such low-level information does not consider semantics associated with the class.


In this paper, we propose another source of guidance that provides a distinction between the foreground and background cues. 
We propose to use the \textit{out-of-distribution (OoD)} data that do not contain any of the foreground classes of interest.
Examples include the rail-only images for the foreground class ``train'', since classifiers often confuse the rail for the train. 
By subduing the recognition of ``train'' on such rail cues in hard OoDs, models successfully distinguish such confusing cues.


Obtaining such OoDs does not incur a significant amount of additional annotation efforts compared to collecting only the image-level labels.
The OoD images are natural by-products of the typical dataset collection procedure. Vision datasets with image-level category labels (\eg Pascal \cite{everingham2010pascal}, COCO~\cite{lin2014microsoft}, LVIS \cite{gupta2019lvis}, and OpenImages \cite{kuznetsova2020open}) all start with a pool of candidate images, from which images corresponding to one of the foreground classes are selected and included in the final dataset. The remaining pool, or the \textit{candidate OoD set}, can be utilized as the source of OoD images. 

The candidate OoD set cannot be directly used for guiding the WSSS method for two reasons. First, general OoD images do not provide informative signals to distinguish difficult background cues from the foreground (\eg rail from train). Second, it may still contain foreground objects. 
We address the first problem by selecting \textit{hard OoDs} whereby classifiers falsely assign high prediction scores to one of the foreground classes.
The second problem is addressed by a human-in-the-loop process where images containing foreground objects are manually pruned. While this requires additional human efforts, we emphasize that the extra cost is negligible. As we will show later (Sec.~\ref{analysis_num_ood_image}), we only need a tiny amount of hard OoD samples to improve the localization maps: even 1 hard OoD image per class boosts the localization performance by 2.0\%p. 
Furthermore, the cost for collecting OoD samples is at the same order of magnitude as collecting the category labels for the foreground samples, as opposed to collecting \eg segmentation maps. 
One can also re-direct the budget for collecting a few labeled foreground data to collecting a similar number of hard OoD samples to dramatically improve the WSSS performance.

Given the additional guidance provided by OoD samples, we propose \textbf{W-OoD}, a method of training a classifier by utilizing the hard-OoDs. 
Note that our data collection procedure provides hard OoD samples which have different patterns and semantics in various contexts.
One could ignore this diversity and treat every hard OoD as a combined background class; this approach has proved to be sub-optimal by our experiments.
Instead, W-OoD considers every hard OoD sample with a metric-learning objective: increase the distance between the in-distribution and OoD samples in the feature space.
This forces the background cues shared by the in-distribution and OoD samples (\eg rail for train category) to be excluded from the feature-space representation. 
W-OoD results in high-quality localization maps and lead to the new state-of-the-art performance on the Pascal VOC 2012 benchmark for WSSS.






We contribute (1) a new paradigm of utilizing the OoD samples to address the spurious correlations in weakly supervised semantic segmentation (WSSS); (2) a dataset of hard OoDs for 20 Pascal categories that will be published upon acceptance; and (3) a WSSS method, W-OoD, that exploits the hard OoDs and achieves the best-known performance on the Pascal VOC 2012 benchmark for WSSS.

\section{Related Work}

\textbf{Weakly supervised learning:}
Most weakly supervised learning methods with image-level class labels are based on a class activation map (CAM)~\cite{zhou2016learning}. However, it is widely
known that a CAM is limited to identifying small discriminative parts of a target object~\cite{lee2019ficklenet, ahn2019weakly, lee2021reducing}.
Several techniques have been proposed for obtaining the entire region of the target object.
PSA~\cite{ahn2018learning} and IRN~\cite{ahn2019weakly} consider pixel relationships to extend the object region to semantically similar areas using a random walk.
SEAM~\cite{wang2020self} regularizes the classifier so that the localization maps obtained from differently transformed images are equivariant to those transformations.
AdvCAM~\cite{lee2021anti} and RIB~\cite{lee2021reducing} propose post-processing techniques of a trained classifier to obtain whole regions of the target object, by manipulating images or network weights.
Although the identified regions are successfully extended by these methods, some spuriously correlated background regions tend to be erroneously identified together. 
CDA~\cite{su2021context} adopts the cut-paste method to decouple the correlation between objects and their contextual background.
However, it is difficult to accurately decouple the correlation using only class labels, which limits the performance improvement. 

\vspace{0.2em}
\textbf{Learning with external data:}
Several studies have considered utilizing additional external information to address the issue of the spurious correlation problem.
Automated web searches can provide images~\cite{shen2018bootstrapping, jin2017webly} or
videos~\cite{hong2017weakly, lee2019frame} with class labels, although these labels may be inaccurate.
Some methods~\cite{li2019attention, sun2020mining} utilize single-label images to obtain more information about in-distribution data.
However, these additional sources still depend solely on classes of interest. Thus, they lack information about the separation between the foreground and background.
Consequently, various types of additional supervision have been adopted.
Some researchers~\cite{vilar2021extracting, sawatzky2019harvesting} employed image captions.
However, these are expensive to obtain. Moreover, modeling vision-language relationships, which is required in those methods, is a non-trivial task.
Kolesnikov \textit{et al.}~\cite{kolesnikov2016improving} proposed an active learning approach, wherein a person determines whether a specific pattern is in the foreground or not.
This is a model-specific approach, so human intervention is required whenever a new model is trained.
Saliency supervision~\cite{wang2017learning, cheng2014global} is another popular additional information source~\cite{lee2021reducing, lee2019frame, sun2020mining, wu2021embedded, yao2021nonsalient, lee2021railroad, joon2017exploiting}. 
However, it is not very effective for non-salient objects that are indistinguishable from the background or small objects~\cite{lee2021reducing, wu2021embedded, lee2021railroad}.


\section{Method}

We propose a method for collecting and utilizing OoD data for the WSSS with category labels. 
We describe the data collection procedure for hard OoD in Sec.~\ref{method_data}. In Sec. \ref{method_proposed}, we introduce the method named W-OoD that trains a classifier with the collected hard-OoDs to generate the localization maps. 
Finally in Sec. \ref{method_trainseg}, we show how to train a semantic segmentation network with the localization maps. 

\subsection{Collecting the Hard OoD Data}\label{method_data}

\begin{figure}[t]
\centering
\includegraphics[width=\linewidth]{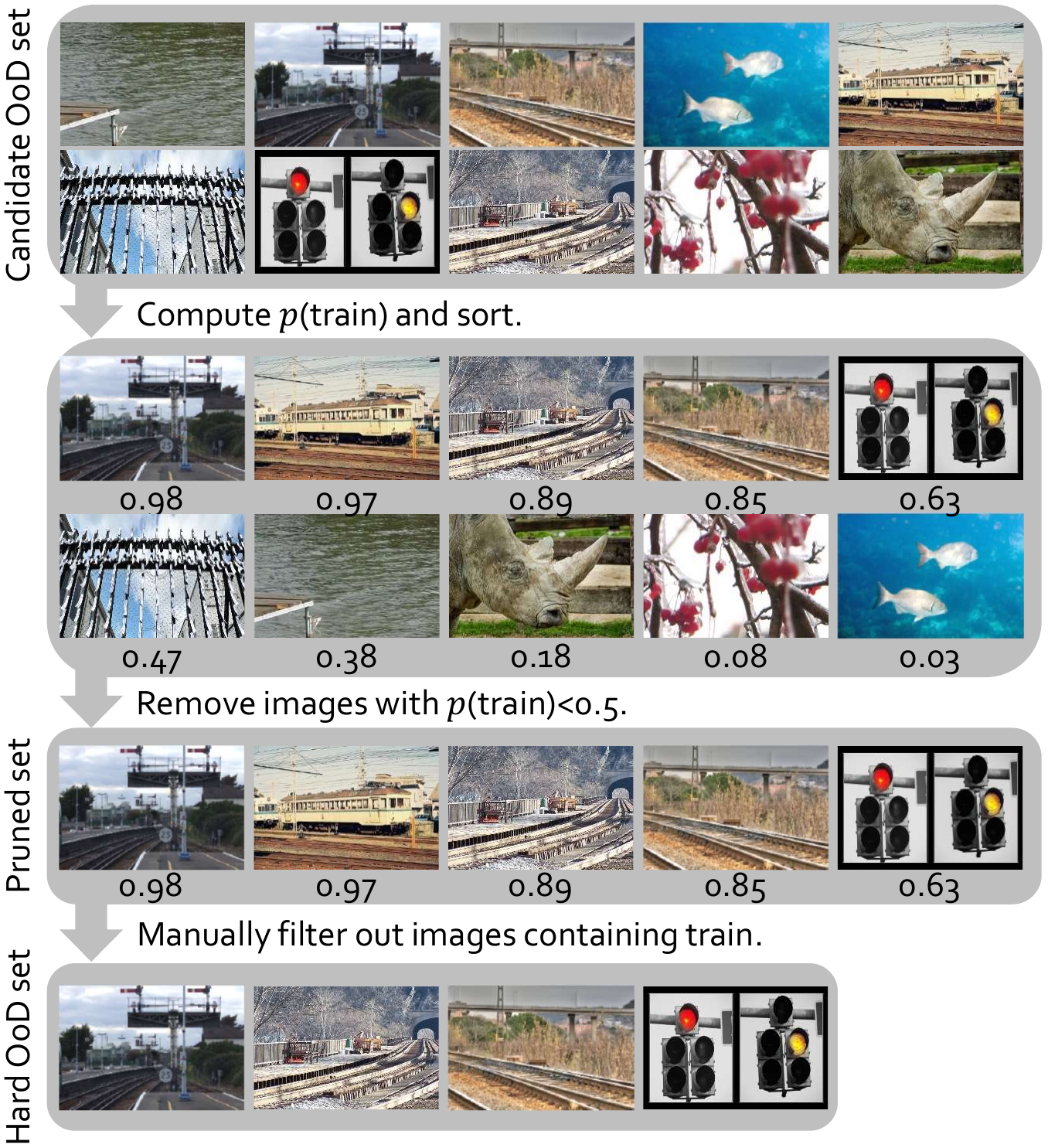}
\vspace{-2em}
\caption{\label{data_collect} \textbf{Collecting hard OoD data}. Starting from the candidate OoD images at the top, we sequentially prune out easy OoDs and then false negatives for each foreground class $c\in\mathcal{C}$. The procedure results in the \textbf{hard OoD dataset}.}
\vspace{-1em}
\end{figure}

We describe the overall procedure for collecting an OoD dataset. The starting point is a \textit{candidate OoD set} that consists mostly of images without the foreground categories of interest. 
The aim is to refine this set into a set of hard OoDs that will be used for the downstream WSSS methods. 
The overall procedure is described in Fig.~\ref{data_collect}.

\textbf{Where to get the candidate OoDs:}
The WSSS task with category labels as the weak supervision first requires the category labels on a set of training images. 
Building a category-labeled image dataset is typically a four-step process \cite{everingham2010pascal, kuznetsova2020open, lin2014microsoft, gupta2019lvis}: (1) define the list $\mathcal{C}$ of foreground classes of interest, (2) acquire unlabelled images from various sources (\eg world wide web), (3) determine for each image whether it contains one of the foreground classes, and (4) tag each image with the foreground category labels.
Steps (3) and (4) are combined in some cases.
A by-product of this procedure is the set of candidate images obtained from step (2) but not selected in step (3). We refer to this set as the \textit{candidate OoD set}. 
For example, for Pascal VOC 2007~\cite{everingham2010pascal}, step (2) has yielded 44,269 candidate images for annotation.
Everingham \textit{et al.}~\cite{everingham2010pascal} report that 9,963 of them were finally selected as foreground data, while the rest were discarded.
We make use of this discarded set that is likely to consist of background images.

\textbf{Hard OoD samples via ranking and pruning:}
Unfortunately, the candidate OoD data are imperfect. OoD data are often too diverse to contain meaningful information. For example, presenting an image of fish in an aquarium as a negative sample of the foreground class ``train'' will not introduce any meaningful supervision for the classifier (See fish in Fig. \ref{data_collect}). It is the \textit{hard OoD samples} that give much information; they are OoD samples confused by a classifier to be containing the foreground object. The rail images \textit{without train} in Fig. \ref{data_collect} are examples of such. They provide informative negative supervision for the classifier to suppress the class score on spurious background cues. 
We thus rank the candidate OoD data according to the prediction scores $p(c)$ for the class $c$ of interest. 
We use the classifier trained on the images with foreground objects and the corresponding labels.
We prune OoD samples with $p(c)<0.5$. This returns candidates for the hard OoD data.



\textbf{Manual pruning of positive samples:}
It is unrealistic to assume that the candidate OoD set will be free of foreground objects. There will be many missing annotations and corner cases. When they are ranked according to the foreground prediction scores, high-ranking images are likely to contain those missing positives. 
We thus need to manually filter out those positive samples.
This manual refinement stage is the cost bottleneck in our pipeline.
The cost depends directly on the \textit{positive rate} $r$, the proportion of positive images among the pruned set obtained by thresholding the prediction score $p(c)\geq 0.5$. 
Letting $n$ be the required number of hard OoD images, the human worker needs to check on average $\frac{n}{1-r}$ images. If there are some positive images with \eg $r=0.2$, then the annotator needs to check $1.25n$ images to eventually obtain $n$ hard OoDs. We denote the resulting dataset as $\mathcal{D}_{\text{ood}}$, the \textit{hard OoD set}.

\textbf{Surrogate source of OoD data:}
Theoretically speaking, it would be best to obtain the hard OoD set by replicating the dataset construction procedure for Pascal \cite{everingham2010pascal} to analyze and benchmark our method on Pascal. 
However, this is practically infeasible because one cannot crawl images with similar characteristics as the 500,000 initial images that Pascal authors have crawled from Flickr in 2007 \cite{everingham2010pascal}. It is also not documented which category annotation tool has been used to filter out the background set.
Another way to set up the experiment is to build a new dataset from scratch.
However, this will not allow us to use the existing WSSS benchmarks like Pascal. 
In this paper, we source the candidate OoD data from another vision dataset: OpenImages \cite{kuznetsova2020open}. 
To simulate the OoD data, we filter out 20 Pascal classes from the OpenImages dataset using the provided category labels.
Note that OpenImages category labels are noisy: 19,794 categories are labeled first through image classifiers and then are refined by crowdsourced workers \cite{kuznetsova2020open}.
This is in stark contrast to Pascal: only 20 categories are labeled by a highly controlled pool of workers at a controlled offline event (called ``annotation party'') \cite{everingham2010pascal}. 
We thus expect the candidate OoD set sourced from OpenImages to contain more noise (\ie foreground classes) than the set one would get from the original Pascal data collection process.

\subsection{Learning with Hard OoD Dataset}\label{method_proposed}
Classifiers trained only on the in-distribution dataset $\mathcal{D}_\text{in}$ often incorrectly identify spuriously correlated background regions as class-relevant patterns.
We address this by using the hard out-of-distribution data $\mathcal{D}_{\text{ood}}$ obtained in the previous section. 
One naive approach to utilize the hard OoD images is either to assign the uniform distribution over the labels for such images (no-information prior)~\cite{hendrycks2018deep, lee2017training, lee2021removing} or to assign the ``background'' label to such images.
However, since hard OoD images contain various semantics that convey meaningful information to each class, labeling these images with one background class ignores the diversity of OoD samples, resulting in a sub-optimal performance as shown in Sec.~\ref{exp_discuss} and Table~\ref{loss_ablation}.

To benefit from the diversity of hard-OoD images, we propose a metric-learning methodology that considers OoD images of individuals or small groups.
To compute a metric-learning objective, we use the penultimate feature $z$ of the in-distribution classifier $\mathcal{F}_\text{in}$ for an input $x$; we write $z_\text{in}$ (resp. $z_\text{ood}$) as the feature of $x_\text{in} \in \mathcal{D}_\text{in}$ (resp. $x_\text{ood} \in \mathcal{D}_\text{ood}$).
We train a classifier $\mathcal{F}$ to ensure that $z_\text{in}$ is significantly different from $z_\text{ood}$, thereby preventing information overlap between the features.
To realize this, a clustering-based metric learning objective is proposed. 

Let $\mathcal{Z}_\text{in}$ and $\mathcal{Z}_\text{ood}$ be the sets of $z_\text{in}$ and $z_\text{ood}$, respectively.
We first construct a set of clusters $\mathcal{P}^\text{in}$ (resp. $\mathcal{P}^\text{ood}$) based on $\mathcal{Z}_\text{in}$ (resp. $\mathcal{Z}_\text{ood}$). 
Each cluster in $\mathcal{P}^\text{in}$ contains features of $x_\text{in}$ corresponding to each class $c\in\mathcal{C}$, resulting in $|\mathcal{C}|$ clusters in $\mathcal{P}^\text{in}$.
One straightforward way of constructing $\mathcal{P}^\text{ood}$ is to cluster images according to their incorrectly predicted classes.
This, however, is sub-optimal in practice because such clusters are highly heterogeneous. For example, images of lakes and images of trees are semantically different, yet a cluster based on the ``bird'' class will contain both. 
Therefore, we construct $\mathcal{P}^\text{ood}$ by using a $K$-means clustering algorithm on $\mathcal{Z}_\text{ood}$.




We now have a set of clusters $\mathcal{P}^\text{in}=\{\mathcal{P}^\text{in}_c\}_{c=1}^{|\mathcal{C}|}$ and $\mathcal{P}^\text{ood}=\{\mathcal{P}^\text{ood}_k\}_{k=1}^{K}$.
The center of each cluster is computed using $p_k = {1 \over |\mathcal{P}_k|} \sum_{x \in \mathcal{P}_k} z(x)$.
We define the distance between the input image $x$ and each cluster $\mathcal{P}_k$ as the distance between $x$'s feature $z(x)$ and the center $p_k$, as follows:
\begin{align}\label{distance}
d(x, \mathcal{P}_k) = \left\lVert z(x) - p_k \right\rVert_2 ~~~ (1\leq k \leq K).
\end{align}
We design a loss $\mathcal{L}_\text{d}$ to ensure that the distance between $x_\text{in}$ and in-distribution clusters $\mathcal{P}^\text{in}$ is small, but the distance between $x_\text{in}$ and OoD clusters $\mathcal{P}^\text{ood}$ is large, as shown below:
\begin{align}\label{distance}
\mathcal{L}_\text{d} = \sum_{c: y_c=1} d(x_\text{in}, \mathcal{P}_c^\text{in}) - \sum_{k \in \mathcal{K}} d(x_\text{in}, \mathcal{P}_k^\text{ood}),
\end{align}
where $y\in\{0,1\}^{|\mathcal{C}|}$ is the multi-hot binary vector of foreground classes in image $x_\text{in}$ and $\mathcal{K}$ is the set of clusters in $\mathcal{P}^\text{ood}$
that are among the top-$\tau \%$ closest from  $x_\text{in}$. This restriction of $\mathcal{K}$ ensures meaningful supervisory signals for the model.



We also use the usual classification loss $\mathcal{L}_\text{cls}$. For in-distribution samples $x_\text{in}$, we use the binary cross entropy (BCE) losses against the label vector $y$. For out-of-distribution samples $x_\text{ood}$, we use the same loss with the zero-vector label $y=(0,\cdots,0)$.
The classification loss for our classifier $\mathcal{F}$ is then
\begin{equation}
    \mathcal{L}_\text{cls} =
    \frac{1}{|\mathcal{C}|}
    \sum_{c=1}^{|\mathcal{C}|}
    \left[
    \mathcal{L}_\text{BCE}(\mathcal{F}^c(x_\text{in}), y_c) + \mathcal{L}_\text{BCE}(\mathcal{F}^c(x_\text{ood}), 0)
    \right],
    \vspace{-0.15em}
\end{equation}
where $\mathcal{F}^c$ is the prediction for class $c$.
The final loss $\mathcal{L}$ to train a classifier $\mathcal{F}$ is
\begin{equation}\label{total_loss}
\vspace{-0.5em}
\mathcal{L} = \mathcal{L}_\text{cls} + \lambda \mathcal{L}_\text{d},
\vspace{-0.00em}
\end{equation}
where $\lambda>0$ is a scalar balancing the two losses.

Because our method adds an additional regularization $\mathcal{L}_\text{d}$ to the existing classifier training, it can be seamlessly integrated into other methods, such as IRN~\cite{ahn2019weakly}, SEAM~\cite{wang2020self} and AdvCAM~\cite{lee2021anti}.


\subsection{Training Segmentation Networks}\label{method_trainseg}
The classifier $\mathcal{F}$ trained by Eq.~\ref{total_loss} generates a localization map using the CAM~\cite{zhou2016learning} technique.
Since the naive CAM generates low-resolution score maps and provides only rough localization of objects, recent WSSS methods~\cite{lee2019ficklenet, wang2020self, lee2021reducing, su2021context, lee2021anti, zhang2020causal} have proposed a framework for expanding the CAM score map to full resolution. They consider the CAM localization map as an initial seed and generate pseudo-ground-truth masks by refining their initial seeds with established seed refinement methods~\cite{huang2018weakly, ahn2018learning, ahn2019weakly, kolesnikov2016seed}. 
In this work, we apply the IRN framework~\cite{ahn2019weakly} on our localization maps to obtain the pseudo-ground-truth masks. They are subsequently used for training segmentation networks.

\begin{table}[tbp]
\renewcommand{\arraystretch}{0.92}
  \centering
  \caption{\textbf{W-OoD improves initial seeds.} We evaluate the qualities of various initial seeds and the effects of applygin W-OoD on them. 
  Evaluated on Pascal VOC 2012 \textit{train} set. All numbers are based on our re-implementation using the official codes.}
  \vspace{-0.7em}
    \begin{tabular}{l@{\hskip 0.07in}cc@{\hskip 0.12in}c@{\hskip 0.07in}c}
     \Xhline{1pt}\\[-0.95em]
    Method  & mIoU  & Prec. & Recall & F1-score \\
    \hline\hline \\[-0.9em]
    $\text{IRN}_{\text{~~CVPR '19}}$~\cite{ahn2018learning} & 49.5  & 61.9 & 72.7  & 66.9\\
    + W-OoD  & \textbf{53.3}  & \textbf{66.5} & \textbf{73.2} & \textbf{69.7}  \\
    \hline \\[-0.9em]
    $\text{SEAM}_{\text{~~CVPR '20}}$~\cite{wang2020self} &  54.8 & 67.2 & 76.5 &  71.5 \\
    + W-OoD  & \textbf{55.9} & \textbf{68.5} & \textbf{76.7} &   \textbf{72.4} \\
    \hline \\[-0.9em]
    $\text{AdvCAM}_{\text{~~CVPR '21}}$~\cite{lee2021anti} & 55.5  & 66.8 & 77.6 & 71.8  \\
    + W-OoD  & \textbf{59.1} & \textbf{71.5} & \textbf{77.9} & \textbf{74.6}  \\
    \Xhline{1pt}
    \vspace{-1.8em}
    \end{tabular}%
  \label{seed_improve}%
\end{table}%

\begin{table}[tbp]
\renewcommand{\arraystretch}{0.95}
  \centering
  \caption{\textbf{Quality of pseudo-GT masks.} mIoU (\%) of the initial seed (Seed), the seed with CRF (+CRF), and the pseudo ground truth mask (Mask) are evaluated on Pascal VOC 2012 \textit{train} set. All the methods based based on IRN~\cite{ahn2019weakly} with ResNet-50.}
  \vspace{-0.7em}
    \begin{tabular}{l@{\hskip 0.3in}c@{\hskip 0.1in}c@{\hskip 0.08in}c}
     \Xhline{1pt}\\[-0.95em]
    Method  &  Seed  & + CRF & Mask \\
    \hline\hline \\[-0.9em]
    
    $\text{IRN}_{\text{~~CVPR '19}}$~\cite{ahn2019weakly} & 49.5  & 54.3 & 66.3 \\
    $\text{MBMNet}_{\text{~~ACMMM '20}}$~\cite{liu2020weakly} & 50.2 & - & 66.8 \\
    $\text{CONTA}_{\text{~~NeurIPS '20}}$~\cite{zhang2020causal} & 48.8 & - & 67.9 \\
    $\text{CDA}_{\text{~~ICCV '21}}$~\cite{su2021context} & 50.8  & - & 67.7 \\
    $\text{AdvCAM}_{\text{~~CVPR '21}}$~\cite{lee2021anti} & 55.6  & 62.1 & 69.9 \\
    $\text{CSE}_{\text{~~ICCV '21}}$~\cite{kweon2021unlocking} & 56.0  & 62.8 & - \\

        \\[-0.9em]
\hline
    \\[-0.9em]
    IRN + W-OoD (Ours)  & 53.3 & 58.4 & 71.1  \\
    AdvCAM + W-OoD (Ours)  & \textbf{59.1} & \textbf{65.5} &\textbf{72.1}  \\

    \Xhline{1pt}
    \vspace{-2em}
    \end{tabular}%
  \label{table_seed}%
\end{table}%

\section{Experiments}
\subsection{Experimental Setup}\label{setup_sec}
\textbf{In-Distribution Dataset:} We conduct experiments on the Pascal VOC 2012~\cite{everingham2010pascal} dataset.
Following the practice in weakly-supervised semantic segmentation (WSSS)~\cite{lee2021anti, ahn2019weakly, wang2020self}, we use the augmented training set containing 10,582 training images produced by Hariharan et al.~\cite{hariharan2011semantic}.
For those training images, we only use the image-level category labels, following the protocol for WSSS.
We use the pixel-wise ground-truth masks on \textit{val} (1,449 images) and \textit{test} (1,456 images) sets only for evaluation.
We use the official Pascal VOC evaluation server for the \textit{test}-set evaluation.

\textbf{Out-of-Distribution Dataset:}
As described in Sec.~\ref{method_data}, we use the OpenImages~\cite{kuznetsova2020open} dataset to construct the candidate OoD set. As the result of prediction-score pruning and manual filtering, we obtain the hard OoD set $\mathcal{D}_\text{ood}$ with 5,190 images.
Examples are shown in the Appendix.


\textbf{Reproducibility:}
We follow experimental settings of IRN~\cite{ahn2019weakly} for training a classifier and obtaining the initial seed, including the use of ResNet-50~\cite{he2016deep}. 
For the setting defined in Sec.~\ref{method_proposed}, we use $\lambda=0.007$, $\tau=20$, and $K=50$. 
For training a segmentation network, we use DeepLab-v2~\cite{chen2017deeplab} with two choices of backbones, ResNet-101~\cite{he2016deep} and Wide ResNet-38~\cite{wu2019wider}, following the practice in recent papers.
All the backbones are pre-trained on ImageNet~\cite{deng2009imagenet}, following existing work~\cite{ahn2018learning, wang2020self, zhang2020causal, kweon2021unlocking, li2021pseudo}.


\begin{figure*}[t]
\centering
\includegraphics[width=0.97\linewidth]{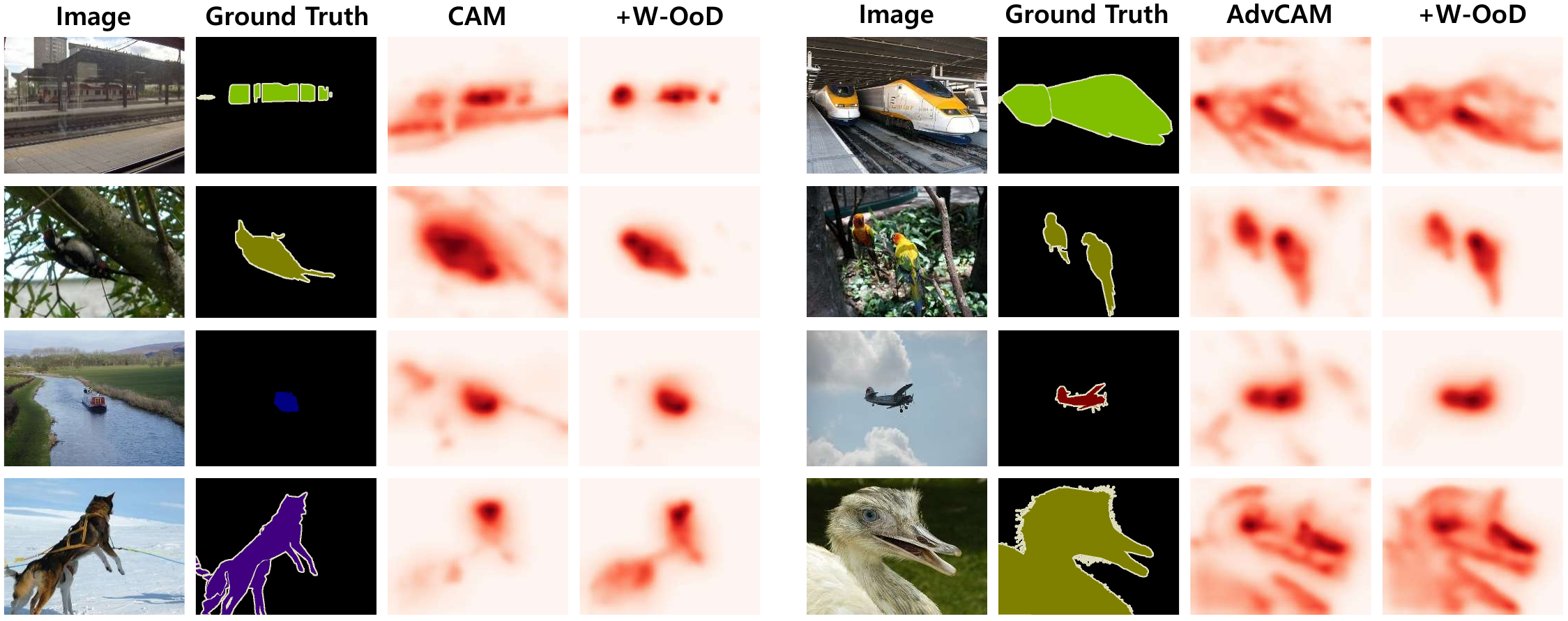}
\vspace{-1em}
\caption{\label{cam_samples} \textbf{Examples of localization maps.} The localization maps are obtained from CAM (left) and AdvCAM~\cite{lee2021anti} (right). In each case, we show the results using our W-OoD method on top.
}
\vspace{-0.5em}
\end{figure*}

\begin{figure*}[t]
\centering
\includegraphics[width=0.97\linewidth]{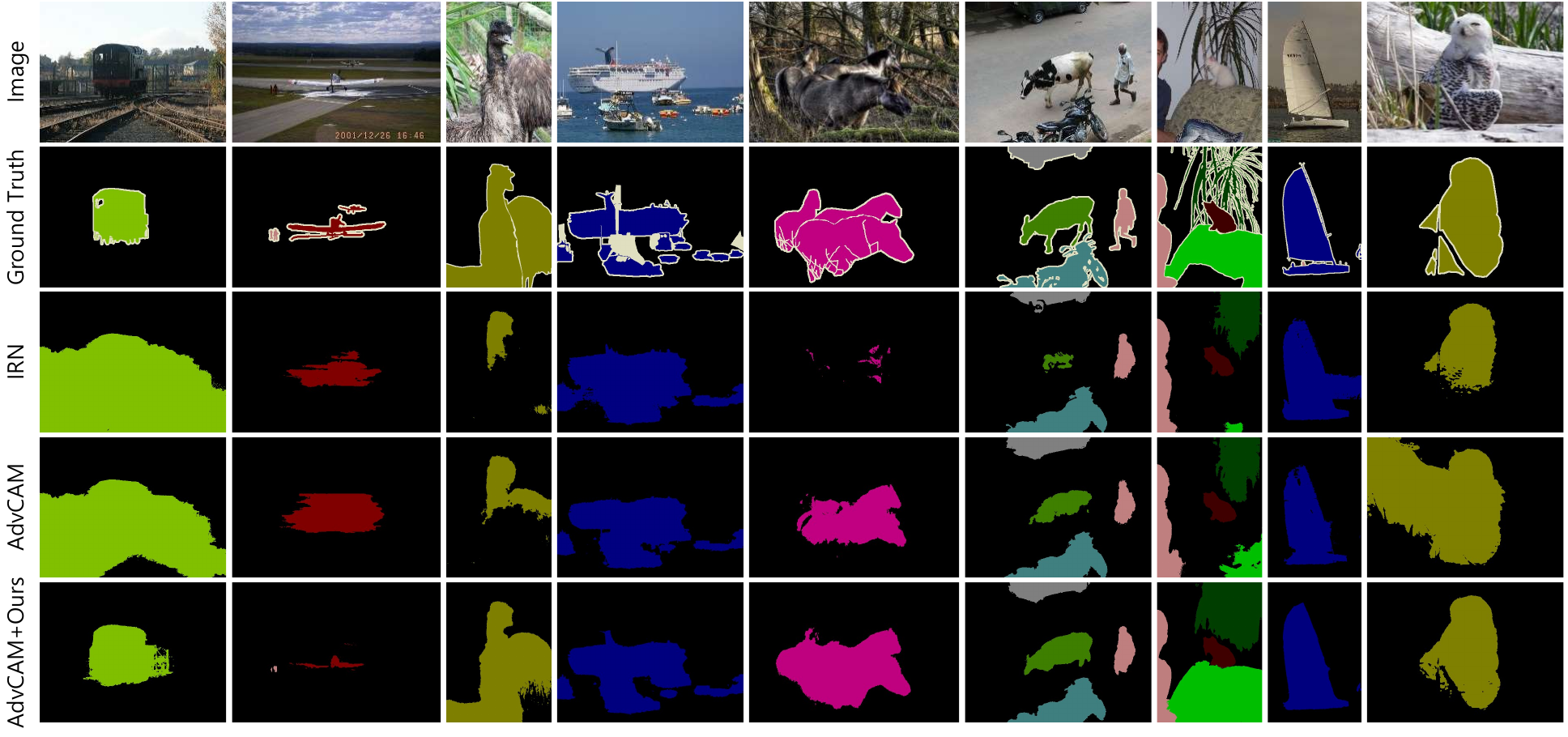}
\vspace{-1.2em}
\caption{\label{segsample} \textbf{Examples of final segmentation results.} Examples of semantic segmentation results on {Pascal VOC 2012} \textit{val} set for IRN~\cite{ahn2019weakly}, AdvCAM~\cite{lee2021anti}, and AdvCAM+Ours. 
}
\vspace{-1em}
\end{figure*}

\subsection{Experimental Results}
\begin{table}[t]
\renewcommand{\arraystretch}{0.95}
\centering
  \caption{\textbf{WSSS performance on Pascal.} We show results on Pascal VOC 2012 \textit{val} and \textit{test} sets. WResNet denotes Wide ResNet~\cite{wu2019wider}. Asterisks $^*$ denote reproduced numbers by us.}\label{table_semantic}
\vspace{-0.7em}
\begin{threeparttable}
\begin{tabular}{l@{\hskip 0.1in}c@{\hskip 0.1in}cc}
    \Xhline{1pt}\\[-0.95em]
    Method  & Backbone& \textit{val} & \textit{test}\\
    \hline\hline 
    \\[-0.9em]
    
    \multicolumn{3}{l}{Supervision: Image-level tags + Saliency}\\
    $\text{FickleNet}_{\text{~~CVPR '19}}$~\cite{lee2019ficklenet}  & ResNet-101 & 64.9 & 65.3\\
    $\text{Sun \textit{et al.}}_{\text{~~ECCV '20}}$~\cite{sun2020mining}   & ResNet-101  & 66.2  & 66.9  \\
    $\text{Yao \textit{et al.}}_{\text{~~CVPR '21}}$~\cite{yao2021nonsalient}   & ResNet-101  & 68.3  & 68.5  \\
    $\text{A$^2$GNN}_{\text{~~TPAMI '21}}$~\cite{zhang2021affinity}   & ResNet-101  &   68.3 & 68.7\\
    $\text{AuxSegNet}_{\text{~~ICCV '21}}$~\cite{xu2021leveraging}    & WResNet-38 & 69.0  & 68.6 \\
    $\text{EDAM}_{\text{~~CVPR '21}}$~\cite{wu2021embedded}    & ResNet-101 & 70.9  & 70.6 \\
        \\[-0.9em]
\hline
    \\[-0.9em]
     \multicolumn{3}{l}{Supervision: Image-level tags}\\
    $\text{IRN}_{\text{~~CVPR '19}}$~\cite{ahn2019weakly}  &  ResNet-50 & 63.5 & 64.8 \\
    $\text{SSDD}_{\text{~~ICCV '19}}$~\cite{Shimoda_2019_ICCV}    & WResNet-38   & 64.9  & 65.5\\
    $\text{SEAM}_{\text{~~CVPR '20}}$~\cite{wang2020self}    & WResNet-38 & 64.5  & 65.7 \\

    $\text{Chang \textit{et al.}}_{\text{~~CVPR '20}}$~\cite{chang2020weakly}   & ResNet-101  & 66.1  & 65.9\\
    
    $\text{CONTA}_{\text{~~NeurIPS '20}}$~\cite{zhang2020causal}   & WResNet-38  & 66.1  & 66.7  \\

    $\text{AdvCAM}_{\text{~~CVPR '21}}$~\cite{lee2021anti}$^*$ & ResNet-101 & 67.5 & 67.1  \\
    $\text{CSE}_{\text{~~ICCV '21}}$~\cite{kweon2021unlocking} & WResNet-38 & 68.3 & 68.0  \\

    $\text{PMM}_{\text{~~ICCV '21}}$~\cite{li2021pseudo} & WResNet-38 & 68.5 & 69.0  \\
        AdvCAM + W-OoD (Ours) & ResNet-101 &  69.8 &  69.9 \\

    AdvCAM + W-OoD (Ours) & WResNet-38 & \textbf{70.7} & \textbf{70.1}  \\
    \Xhline{1pt}
    
    \end{tabular}%
     \end{threeparttable}
    \vspace{-1em}

      \end{table}

\textbf{Quality of localization maps:}
As mentioned in Sec.~\ref{method_proposed}, our method can be applied to other WSSS methods, since it only requires the addition of a loss term $\mathcal{L}_\text{d}$ during the classifier training.
We apply our method to three state-of-the-art WSSS methods that utilize the initial seeds: IRN~\cite{ahn2019weakly}, SEAM~\cite{wang2020self}, and AdvCAM~\cite{lee2021anti}.
Table~\ref{seed_improve} presents the qualities of the initial seeds for the considered baselines as well as respective performances when combined with our W-OoD technique. We observe that our method improves all the metrics by a large margin for all three methods. 
In particular, W-OoD training significantly improves precision values (\eg +4.7\%p for AdvCAM~\cite{lee2021anti}), indicating that the resulting localization maps bleed into the background regions less frequently. This is what we expected to see as a result of including the hard OoD samples into training.
Fig.~\ref{cam_samples} shows qualitative examples of the localization maps. 
They show that our method generates more precise maps around the actual foreground objects. Spuriously correlated background regions like rails for ``train'' and trees for ``bird'' are effectively suppressed by our method.
Additionally, we observe that our method improves recall by expanding the retrieved region of the target object, as shown in the last column in Fig.~\ref{cam_samples}. 
The increased precision gives room for further improvements in recall.

\begin{table*}[t]
  \centering
\begin{minipage}{0.66\linewidth}
  \centering
\includegraphics[width=\linewidth]{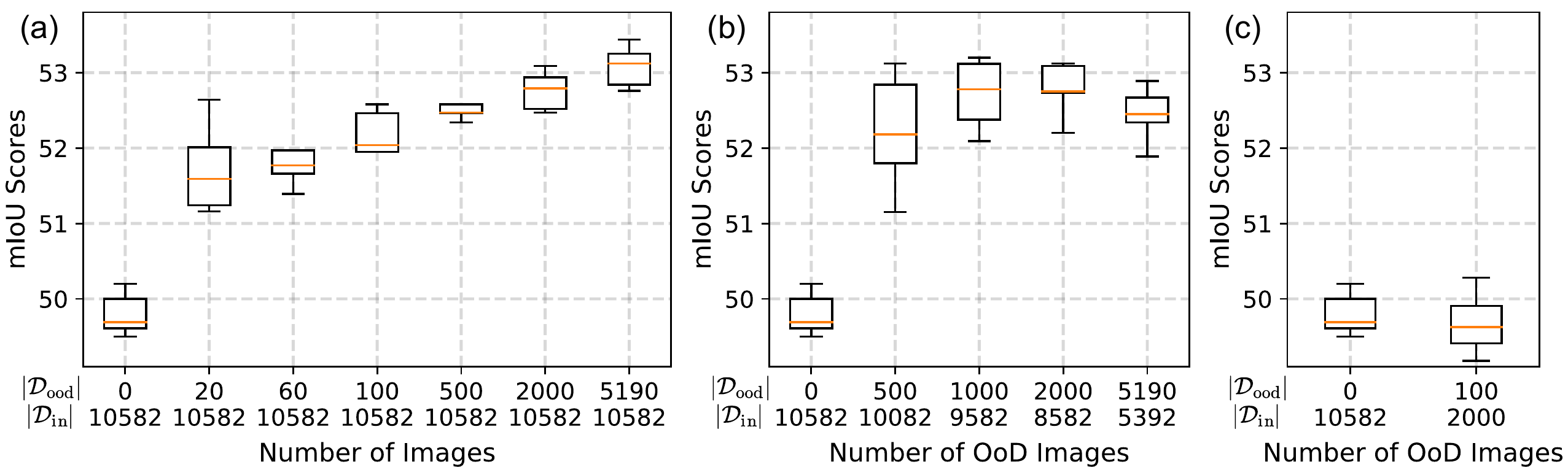}
  \vspace{-2em}
  \captionof{figure}{\label{fig_n_ood} \textbf{Amount of hard OoD samples.} We vary number of in-distribution training data $\mathcal{D}_\text{in}$ (originally 10,582) and the hard OoD data (originally 0). (a) We fix $|\mathcal{D}_\text{in}|=10,582$ and vary $|\mathcal{D}_\text{ood}|$. (b) We fix $|\mathcal{D}_\text{in}|+|\mathcal{D}_\text{ood}|=10,582$ and vary $|\mathcal{D}_\text{ood}|$. (c) We use $|\mathcal{D}_\text{in}|=2,000$ and $|\mathcal{D}_\text{ood}|=100$.
  The box plots show the quantiles over five repeated experiments.}
  \end{minipage}\hfill
\begin{minipage}{0.3\linewidth}
\center
\small
{
\begin{tabular}{ccc}
\Xhline{1pt}\\[-0.95em]
    $K$     & Clustering & mIoU \\
    \hline\hline
       20   &   Predicted classes   &  52.1\\
       \hline
    20      &   \multirow{4}[0]{*}{K-Means}  & 52.4 \\
     30     &       &  53.1\\
      50    &       &  \textbf{53.3}\\

    70      &      & 52.6 \\
    \Xhline{1pt}
    \end{tabular}%
    }
      \caption{\textbf{Constructing $\mathcal{P}^\text{ood}$. }
      We compare two methods for constructing $\mathcal{P}^\text{ood}$ for W-OoD training. 
      We report the mIoU of the initial seeds on Pascal VOC 2012 \textit{train} set.}
  \label{table_clustering}%
  \end{minipage}\hfill
\vspace{-1em}
\end{table*}

\begin{figure*}[t]
\centering
\includegraphics[width=0.95\linewidth]{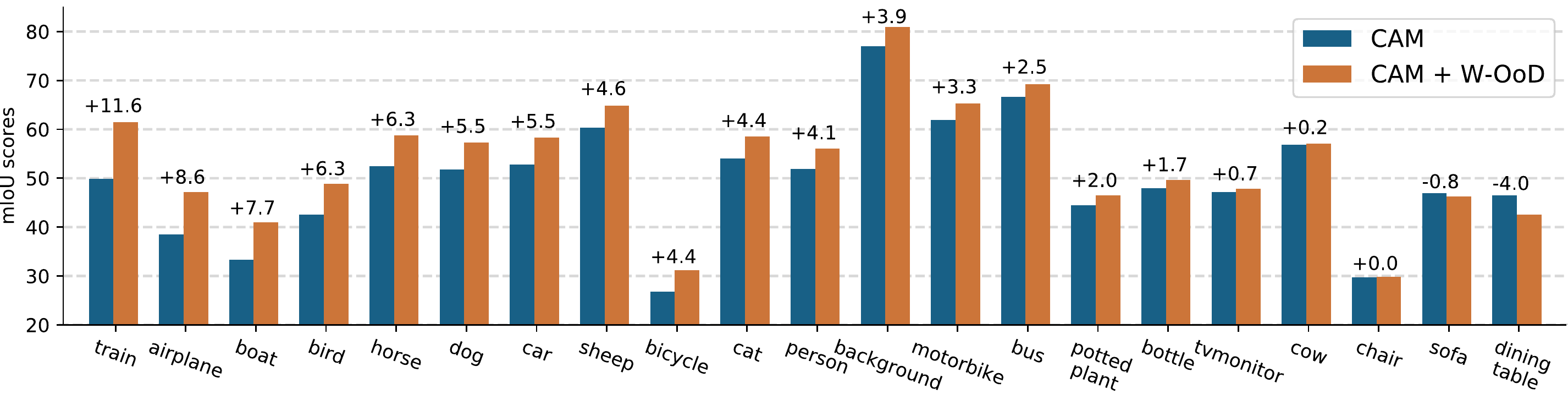}
\vspace{-0.7em}
\caption{\label{class_wise} \textbf{Per-class seed qualities.} We compare the baseline IRN~\cite{ahn2019weakly} (denoted as ``CAM'' above) and the W-OoD augmented version for each class. Evaluated on Pascal VOC 2012 \textit{train} set. Classes are sorted in the descending order of $\Delta$improvement (\%p).}
\vspace{-1em}
\end{figure*}

\textbf{Quality of pseudo-ground-truth masks:}
Table~\ref{table_seed} compares qualities of intermediate masks leading to the pseudo-ground-truth masks among state-of-the-art methods as well as ours.
Our pseudo ground-truth masks achieve an mIoU value of 72.1, which outperforms the previous state of the art by a large margin.
Note that CDA~\cite{su2021context} is likewise motivated by the need to suppress spurious correlations between foreground and background cues, but has only used the in-distribution data to tackle the problem.
It improves the initial seed of IRN~\cite{ahn2019weakly} by 1.3\%p mIoU (49.5 $\rightarrow$ 50.8), while our method improves it by 3.8\%p mIoU (49.5 $\rightarrow$ 53.3, in Table~\ref{seed_improve}).
We believe that in-distribution data are fundamentally limited in providing sufficient evidence for distinguishing certain background cues from foreground: if one always sees train on rail, how can one learn that rail is not part of the train?
We believe this missing knowledge is effectively supplied by the hard OoD images.






\textbf{Final segmentation results:}
We present the WSSS benchmark results in Table~\ref{table_semantic}.
It achieve the best result among the variants using only image-level tags: 70.7\% mIoU on \emph{val} and 70.1\% mIoU on \emph{test}. 
In particular, using the same backbone ResNet-101~\cite{he2016deep}, our method produces 2.3\%p better mIoU than the baseline AdvCAM~\cite{lee2021anti}.
Our method also outperforms other methods using additional saliency supervision~\cite{li2014secrets, liu2010learning} that explicitly provides pixel-level information of salient objects in an image, except for EDAM~\cite{wu2021embedded}.
Fig.~\ref{segsample} shows examples of semantic masks produced by IRN~\cite{ahn2019weakly}, AdvCAM~\cite{lee2021anti}, and our AdvCAM + W-OoD.
In the examples, our method captures the extent of the target objects more precisely than the baselines.


\begin{figure*}[t]
\centering
\includegraphics[width=\linewidth]{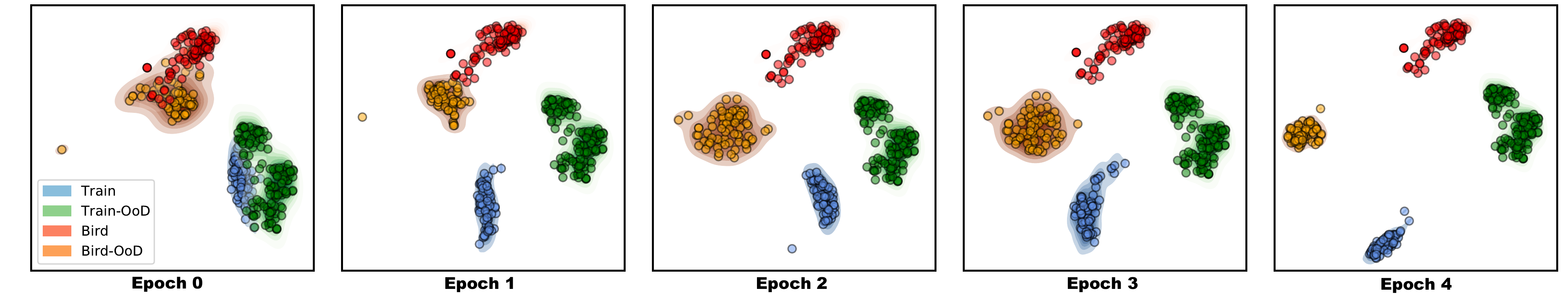}
\vspace{-1.7em}
\caption{\label{tsne_samples} \textbf{Visualization of intermediate features.} We visualize the intermediate features for ``train'' and ``bird'' classes, as well as the features for respective OoD samples, at different training stages. We use the T-SNE~\cite{maaten2008visualizing} dimensionality reduction technique.}
\vspace{-0.7em}
\end{figure*}

\vspace{-0.1em}
\subsection{Analysis and Discussion}\label{exp_discuss}
\vspace{-0.1em}
\subsubsection{Number of OoD Images}\label{analysis_num_ood_image}
\vspace{-0.3em}
We investigate the impact of the number of OoD images for our W-OoD training method. Fig.~\ref{fig_n_ood}(a) shows the mIoU scores of the initial seed at different numbers of OoD images ($|\mathcal{D}_\text{ood}|$) while keeping the number of in-distribution images constant at $|\mathcal{D}_\text{in}|=10,582$. 
The experiments were repeated five times to investigate the sensitivity of the result to different random subsets of $\mathcal{D}_\text{ood}$.
We observe that already at 1 hard OoD sample per class ($|\mathcal{D}_\text{ood}|=20$), the performance boost is 2.0\%p (49.8 $\rightarrow$ 51.8), though with a significant amount of variance.
The marginal gain from additional hard OoD images diminishes with increasing number of samples.
The performance variance also diminishes with an increased number of hard OoD samples.

In the second experiment, we vary the number of hard OoD samples $|\mathcal{D}_\text{ood}|$ while fixing the total number of image-level labeled samples: $|\mathcal{D}_\text{in}| + |\mathcal{D}_\text{ood}| = 10,582$.
This is a version of fixing the budget for in-distribution and out-of-distribution samples.
Fig.~\ref{fig_n_ood}(b) shows that the hard OoD images  bring far greater unit gain than in-distribution images. Thus, given a fixed budget, it is advisable to spend at least some portion of it on collecting the hard OoD samples. 

In Fig.~\ref{fig_n_ood}(c), we observe that, with 100 hard OoD images, we only need 2,000 in-distribution images to match the performance we obtain from the original 10,582 in-distribution images, enhancing the data efficiency by around 500\%.




\vspace{-0.7em}

\subsubsection{Effectiveness of Each Component}
\vspace{-0.3em}

\textbf{K-Means clustering:} Table~\ref{table_clustering} compares the two methods for constructing the $\mathcal{P}^\text{ood}$ in Sec.~\ref{method_proposed}.
When the OoD clusters are based on the classes predicted by the classifier, the resulting mIoU is 52.1\%, which is not significantly different from that obtained using the K-means clustering method for the same $K$ value.
The clustering method based on the predicted class limits $K$ to $|\mathcal{C}|$, whereas $K$ values can be controlled in K-means clustering.
At $K=50$, it produces an mIoU value of 53.3\% and the performance is stable across a broad range of $K$ values. Examples of OoD samples in each cluster are presented in the Appendix.

\begin{table}[t]
\renewcommand{\arraystretch}{0.9}
  \centering
  \small
  \caption{\textbf{Loss ablations.} Effectiveness of each loss on the initial seed in mIoU(\%) on Pascal VOC \textit{train} set.}
  \vspace{-1em}
    \begin{tabular}{cccccccc}
    \Xhline{1pt}
    Loss & Data &(a)&(b)&(c)&(d)&(e)&(f)\\\hline
    \multirow{2}[0]{*}{$\mathcal{L}_\text{cls}$} & $\mathcal{D}_\text{in}$   &   \textcolor{black}{\cmark}    &    \textcolor{black}{\cmark}   &    \textcolor{black}{\cmark}   &  \textcolor{black}{\cmark}  & \textcolor{black}{\cmark}   & \textcolor{black}{\cmark}  \\
          & $\mathcal{D}_\text{ood}$   &       &  \textcolor{black}{\cmark}     &       &    \textcolor{black}{\cmark}  & \textcolor{black}{\cmark} &  \textcolor{black}{\cmark} \\
    \hline
    \multirow{2}[0]{*}{$\mathcal{L}_\text{d}$} & $\mathcal{D}_\text{in}$   &       &       &   \textcolor{black}{\cmark}    &     \textcolor{black}{\cmark}  &  &\textcolor{black}{\cmark} \\
          & $\mathcal{D}_\text{ood}$   &       &       &    \textcolor{black}{\cmark}   &       & \textcolor{black}{\cmark}& \textcolor{black}{\cmark}  \\
    \hline\hline \\[-0.9em]
    \multicolumn{2}{c}{mIoU} & 49.5  &  50.0     &  52.5    &    50.2   &52.3&  53.3 \\
    \Xhline{1pt}
    \vspace{-2em}
    \end{tabular}%
  \label{loss_ablation}%
\end{table}%

\noindent\textbf{Loss functions:} 
We conduct ablation studies for each loss in Eq.~\ref{total_loss}. 
Both $\mathcal{L}_\text{cls}$ and $\mathcal{L}_\text{d}$ consist of terms for in-distribution $\mathcal{D}_\text{in}$ and out-of-distribution $\mathcal{D}_\text{ood}$ data. The effectiveness of each loss term as well as the dataset type is presented in Table~\ref{loss_ablation}. (a) is the result of using only $\mathcal{L}_\text{cls}$ for $\mathcal{D}_\text{in}$, which is our baseline. 
The performance boost for (a)$\rightarrow$(b) and (c)$\rightarrow$(e) indicates that training the classifier to predict OoD images as background ($\mathcal{L}_\text{cls}$ on $\mathcal{D}_\text{ood}$) is effective, though with only marginal improvements.
The improvement along (b)$\rightarrow$(d)$\rightarrow$(f) signifies the importance of $\mathcal{L}_\text{d}$, in particular when used on the hard OoD data $\mathcal{D}_\text{ood}$. We also find that $\mathcal{L}_\text{d}$ for $\mathcal{D}_\text{in}$ is useful for stabilizing the performance: in (e)$\rightarrow$(f), the standard deviation decreases from 0.82 to 0.33.

\vspace{-1em}
\subsubsection{Analysis of Results by Class}
\vspace{-0.2em}
Different object classes exhibit different amounts of spurious correlation with background. 
For example, ``train'' objects are often confused with the rail background due to their high co-occurrence with rails. 
Objects like ``tvmonitor'', on the other hand, suffer less from this issue because of the variety of the co-occuring concepts: a TV can be freely put next to a wall, furniture, window, or any other indoor objects.
We show the class-wise performances for the baseline IRN~\cite{ahn2019weakly} and ours in
Fig.~\ref{class_wise}.
First of all, we note that our method improves the class-wise performances rather proportionately: 18 out of 21 classes have seen a performance improvement.
Classes that have benefited most from our method are train, airplane, boat, bird, and horse. 
They are ones that are well-known for spurious background correlations: 
train-rail, airplane-sky/runway, boat-water, bird-tree/sky, and horse-meadow.

On the other hand, a particularly large drop in mIoU is seen for the ``dining table'' class.
We conjecture the spurious background correlation has actually been helping out the localization of the ``dining table'' objects. 
Many pixel-wise ground-truth evaluation mask for ``dining table'' objects erroneously include the objects put on it, such as plates, cutlery, and foods.
By labeling OoD images, which contain those co-occurring objects not put on a dining table, as ``no dining table'', the model may correctly assign lower ``dining table'' scores on those objects, ironically harming the final performance measured on noisy masks.
See Appendix for the examples.
We believe there will be an additional performance gain if those wrong ground-truth masks are fixed.


\vspace{-0.9em}
\subsubsection{Manifold Visualization}
\vspace{-0.4em}
To observe the training dynamics of our method, we visualize the feature manifold at different stages of the W-OoD training. 
We collect two sets of images with respective labels ``train'' and ``bird'' from $\mathcal{D}_\text{in}$ and two sets of images which are respectively falsely predicted as ``train'' and ``bird'' by $\mathcal{F}_\text{in}$ from $\mathcal{D}_\text{ood}$.
Using the classifier at epoch $e\in\{0,\cdots5\}$\footnote{The classifier at $e=0$ is the one trained using in-distribution images.}, we compute the features $z_\text{in}$ and $z_\text{out}$ from images drawn from $\mathcal{D}_\text{in}$ and $\mathcal{D}_\text{ood}$, respectively.
We use t-SNE~\cite{maaten2008visualizing} to reduce the dimensionality of each feature to 2 dimensions.
Fig.~\ref{tsne_samples} visualizes the features $z_\text{in}$ and $z_\text{out}$ after dimensional reduction using t-SNE. 
It is observed that, at the beginning of the epoch, $z_\text{in}$ and $z_\text{out}$ of each class are rarely distinguishable, indicating that the classifier encodes similar information for in-distribution and OoD images. However, as W-OoD training progresses, the two features gradually become distinct.
This analysis supports the argument that our method allows the classifier to avoid modeling common information between in-distribution and OoD images, as intended.










\vspace{-0.5em}
\section{Conclusion and Future Directions}
\vspace{-0.3em}
We have proposed the use of a new source of information, the OoD data, for suppressing the spurious correlations learned by weakly supervised semantic segmentation (WSSS) methods. 
We have showcased the data collection pipeline whereby the suitable hard OoD images are obtained.
By including those images as negative samples in addition to the original in-distribution foreground samples, we have been able to train a classifier with more accurate localization maps. 
Our method achieves a performance superior to existing WSSS methods based on image-level labels.
In addition, we have empirically shown that the image-level labeling cost itself can be further reduced by using the hard OoD images, without sacrificing the WSSS performances. 
We have focused on using OoD images for training classifiers to produce accurate pseudo ground-truth masks; interesting future work will include exploiting the OoD images in training a segmentation network itself.


\bigskip
\vspace{-0.5em}
\noindent\textbf{Acknowledgements:}
This work was supported by Institute of Information \& communications Technology Planning \& Evaluation (IITP) grant funded by the Korea government (MSIT) [NO.2021-0-01343, Artificial Intelligence Graduate School Program (Seoul National University)], 
AIRS Company in Hyundai Motor and Kia through HMC/KIA-SNU AI Consortium Fund, and the Brain Korea 21 Plus Project in 2021.

{\small
\bibliographystyle{ieee_fullname}
\bibliography{egbib}

\begin{thebibliography}{10}\itemsep=-1pt

\bibitem{ahn2019weakly}
Jiwoon Ahn, Sunghyun Cho, and Suha Kwak.
\newblock Weakly supervised learning of instance segmentation with inter-pixel
  relations.
\newblock In {\em CVPR}, 2019.

\bibitem{ahn2018learning}
Jiwoon Ahn and Suha Kwak.
\newblock Learning pixel-level semantic affinity with image-level supervision
  for weakly supervised semantic segmentation.
\newblock In {\em CVPR}, 2018.

\bibitem{bearman2016s}
Amy Bearman, Olga Russakovsky, Vittorio Ferrari, and Li Fei-Fei.
\newblock What’s the point: Semantic segmentation with point supervision.
\newblock In {\em ECCV}, 2016.

\bibitem{chang2020weakly}
Yu-Ting Chang, Qiaosong Wang, Wei-Chih Hung, Robinson Piramuthu, Yi-Hsuan Tsai,
  and Ming-Hsuan Yang.
\newblock Weakly-supervised semantic segmentation via sub-category exploration.
\newblock In {\em CVPR}, 2020.

\bibitem{chen2017deeplab}
Liang-Chieh Chen, George Papandreou, Iasonas Kokkinos, Kevin Murphy, and Alan~L
  Yuille.
\newblock Deeplab: Semantic image segmentation with deep convolutional nets,
  atrous convolution, and fully connected crfs.
\newblock {\em IEEE TPAMI}, 2017.

\bibitem{cheng2014global}
Ming-Ming Cheng, Niloy~J Mitra, Xiaolei Huang, Philip~HS Torr, and Shi-Min Hu.
\newblock Global contrast based salient region detection.
\newblock {\em TPAMI}, 2014.

\bibitem{choe2020evaluating}
Junsuk Choe, Seong~Joon Oh, Seungho Lee, Sanghyuk Chun, Zeynep Akata, and
  Hyunjung Shim.
\newblock Evaluating weakly supervised object localization methods right.
\newblock In {\em CVPR}, 2020.

\bibitem{mmseg2020}
MMSegmentation Contributors.
\newblock {MMSegmentation}: Openmmlab semantic segmentation toolbox and
  benchmark.
\newblock \url{https://github.com/open-mmlab/mmsegmentation}, 2020.

\bibitem{cordts2016cityscapes}
Marius Cordts, Mohamed Omran, Sebastian Ramos, Timo Rehfeld, Markus Enzweiler,
  Rodrigo Benenson, Uwe Franke, Stefan Roth, and Bernt Schiele.
\newblock The cityscapes dataset for semantic urban scene understanding.
\newblock In {\em CVPR}, 2016.

\bibitem{deng2009imagenet}
Jia Deng, Wei Dong, Richard Socher, Li-Jia Li, Kai Li, and Li Fei-Fei.
\newblock Imagenet: A large-scale hierarchical image database.
\newblock In {\em CVPR}, 2009.

\bibitem{everingham2010pascal}
Mark Everingham, Luc Van~Gool, Christopher~KI Williams, John Winn, and Andrew
  Zisserman.
\newblock The pascal visual object classes (voc) challenge.
\newblock {\em IJCV}, 2010.

\bibitem{fan2018cian}
Junsong Fan, Zhaoxiang Zhang, and Tieniu Tan.
\newblock Cian: Cross-image affinity net for weakly supervised semantic
  segmentation.
\newblock {\em AAAI}, 2020.

\bibitem{gupta2019lvis}
Agrim Gupta, Piotr Dollar, and Ross Girshick.
\newblock {LVIS}: A dataset for large vocabulary instance segmentation.
\newblock In {\em CVPR}, 2019.

\bibitem{hariharan2011semantic}
Bharath Hariharan, Pablo Arbel{\'a}ez, Lubomir Bourdev, Subhransu Maji, and
  Jitendra Malik.
\newblock Semantic contours from inverse detectors.
\newblock In {\em ICCV}, 2011.

\bibitem{he2016deep}
Kaiming He, Xiangyu Zhang, Shaoqing Ren, and Jian Sun.
\newblock Deep residual learning for image recognition.
\newblock In {\em CVPR}, 2016.

\bibitem{hendrycks2018deep}
Dan Hendrycks, Mantas Mazeika, and Thomas Dietterich.
\newblock Deep anomaly detection with outlier exposure.
\newblock In {\em ICLR}, 2019.

\bibitem{hong2017weakly}
Seunghoon Hong, Donghun Yeo, Suha Kwak, Honglak Lee, and Bohyung Han.
\newblock Weakly supervised semantic segmentation using web-crawled videos.
\newblock In {\em CVPR}, 2017.

\bibitem{huang2018weakly}
Zilong Huang, Xinggang Wang, Jiasi Wang, Wenyu Liu, and Jingdong Wang.
\newblock Weakly-supervised semantic segmentation network with deep seeded
  region growing.
\newblock In {\em CVPR}, 2018.

\bibitem{jin2017webly}
Bin Jin, Maria~V Ortiz~Segovia, and Sabine Susstrunk.
\newblock Webly supervised semantic segmentation.
\newblock In {\em CVPR}, 2017.

\bibitem{joon2017exploiting}
Seong Joon~Oh, Rodrigo Benenson, Anna Khoreva, Zeynep Akata, Mario Fritz, and
  Bernt Schiele.
\newblock Exploiting saliency for object segmentation from image level labels.
\newblock In {\em CVPR}, 2017.

\bibitem{ke2021universal}
Tsung-Wei Ke, Jyh-Jing Hwang, and Stella~X Yu.
\newblock Universal weakly supervised segmentation by pixel-to-segment
  contrastive learning.
\newblock In {\em ICLR}, 2021.

\bibitem{khoreva2017simple}
Anna Khoreva, Rodrigo Benenson, Jan Hosang, Matthias Hein, and Bernt Schiele.
\newblock Simple does it: Weakly supervised instance and semantic segmentation.
\newblock In {\em CVPR}, 2017.

\bibitem{kim2021beyond}
Beomyoung Kim, Youngjoon Yoo, Chaeeun Rhee, and Junmo Kim.
\newblock Beyond semantic to instance segmentation: Weakly-supervised instance
  segmentation via semantic knowledge transfer and self-refinement.
\newblock {\em arXiv preprint arXiv:2109.09477}, 2021.

\bibitem{kolesnikov2016improving}
Alexander Kolesnikov and Christoph~H Lampert.
\newblock Improving weakly-supervised object localization by micro-annotation.
\newblock In {\em BMVC}, 2016.

\bibitem{kolesnikov2016seed}
Alexander Kolesnikov and Christoph~H Lampert.
\newblock Seed, expand and constrain: Three principles for weakly-supervised
  image segmentation.
\newblock In {\em ECCV}, 2016.

\bibitem{kuznetsova2020open}
Alina Kuznetsova, Hassan Rom, Neil Alldrin, Jasper Uijlings, Ivan Krasin, Jordi
  Pont-Tuset, Shahab Kamali, Stefan Popov, Matteo Malloci, Alexander
  Kolesnikov, et~al.
\newblock The open images dataset v4.
\newblock {\em IJCV}, 2020.

\bibitem{kwak2017weakly}
Suha Kwak, Seunghoon Hong, and Bohyung Han.
\newblock Weakly supervised semantic segmentation using superpixel pooling
  network.
\newblock In {\em AAAI}, 2017.

\bibitem{kweon2021unlocking}
Hyeokjun Kweon, Sung-Hoon Yoon, Hyeonseong Kim, Daehee Park, and Kuk-Jin Yoon.
\newblock Unlocking the potential of ordinary classifier: Class-specific
  adversarial erasing framework for weakly supervised semantic segmentation.
\newblock In {\em ICCV}, 2021.

\bibitem{lee2021reducing}
Jungbeom Lee, Jooyoung Choi, Jisoo Mok, and Sungroh Yoon.
\newblock Reducing information bottleneck for weakly supervised semantic
  segmentation.
\newblock In {\em NeurIPS}, 2021.

\bibitem{lee2019ficklenet}
Jungbeom Lee, Eunji Kim, Sungmin Lee, Jangho Lee, and Sungroh Yoon.
\newblock Ficklenet: Weakly and semi-supervised semantic image segmentation
  using stochastic inference.
\newblock In {\em CVPR}, 2019.

\bibitem{lee2019frame}
Jungbeom Lee, Eunji Kim, Sungmin Lee, Jangho Lee, and Sungroh Yoon.
\newblock Frame-to-frame aggregation of active regions in web videos for weakly
  supervised semantic segmentation.
\newblock In {\em ICCV}, 2019.

\bibitem{lee2021anti}
Jungbeom Lee, Eunji Kim, and Sungroh Yoon.
\newblock Anti-adversarially manipulated attributions for weakly and
  semi-supervised semantic segmentation.
\newblock In {\em CVPR}, 2021.

\bibitem{lee2021bbam}
Jungbeom Lee, Jihun Yi, Chaehun Shin, and Sungroh Yoon.
\newblock Bbam: Bounding box attribution map for weakly supervised semantic and
  instance segmentation.
\newblock In {\em CVPR}, 2021.

\bibitem{lee2017training}
Kimin Lee, Honglak Lee, Kibok Lee, and Jinwoo Shin.
\newblock Training confidence-calibrated classifiers for detecting
  out-of-distribution samples.
\newblock In {\em ICLR}, 2018.

\bibitem{lee2018robust}
Sungmin Lee, Jangho Lee, Jungbeom Lee, Chul-Kee Park, and Sungroh Yoon.
\newblock Robust tumor localization with pyramid grad-cam.
\newblock {\em arXiv preprint arXiv:1805.11393}, 2018.

\bibitem{lee2021railroad}
Seungho Lee, Minhyun Lee, Jongwuk Lee, and Hyunjung Shim.
\newblock Railroad is not a train: Saliency as pseudo-pixel supervision for
  weakly supervised semantic segmentation.
\newblock In {\em CVPR}, 2021.

\bibitem{lee2021removing}
Saehyung Lee, Changhwa Park, Hyungyu Lee, Jihun Yi, Jonghyun Lee, and Sungroh
  Yoon.
\newblock Removing undesirable feature contributions using out-of-distribution
  data.
\newblock In {\em ICLR}, 2021.

\bibitem{li2019guided}
Kunpeng Li, Ziyan Wu, Kuan-Chuan Peng, Jan Ernst, and Yun Fu.
\newblock Guided attention inference network.
\newblock {\em IEEE TPAMI}, 2019.

\bibitem{li2019attention}
Kunpeng Li, Yulun Zhang, Kai Li, Yuanyuan Li, and Yun Fu.
\newblock Attention bridging network for knowledge transfer.
\newblock In {\em CVPR}, 2019.

\bibitem{li2014secrets}
Yin Li, Xiaodi Hou, Christof Koch, James~M Rehg, and Alan~L Yuille.
\newblock The secrets of salient object segmentation.
\newblock In {\em CVPR}, 2014.

\bibitem{li2021pseudo}
Yi Li, Zhanghui Kuang, Liyang Liu, Yimin Chen, and Wayne Zhang.
\newblock Pseudo-mask matters in weakly-supervised semantic segmentation.
\newblock In {\em ICCV}, 2021.

\bibitem{lin2014microsoft}
Tsung-Yi Lin, Michael Maire, Serge Belongie, James Hays, Pietro Perona, Deva
  Ramanan, Piotr Doll{\'a}r, and C~Lawrence Zitnick.
\newblock Microsoft coco: Common objects in context.
\newblock In {\em ECCV}, 2014.

\bibitem{liu2010learning}
Tie Liu, Zejian Yuan, Jian Sun, Jingdong Wang, Nanning Zheng, Xiaoou Tang, and
  Heung-Yeung Shum.
\newblock Learning to detect a salient object.
\newblock {\em TPAMI}, 2010.

\bibitem{liu2020weakly}
Weide Liu, Chi Zhang, Guosheng Lin, Tzu-Yi HUNG, and Chunyan Miao.
\newblock Weakly supervised segmentation with maximum bipartite graph matching.
\newblock In {\em ACMMM}, 2020.

\bibitem{liu2020leveraging}
Yun Liu, Yu-Huan Wu, Pei-Song Wen, Yu-Jun Shi, Yu Qiu, and Ming-Ming Cheng.
\newblock Leveraging instance-, image-and dataset-level information for weakly
  supervised instance segmentation.
\newblock {\em TPAMI}, 2020.

\bibitem{maaten2008visualizing}
Laurens van~der Maaten and Geoffrey Hinton.
\newblock Visualizing data using t-sne.
\newblock {\em Journal of machine learning research}, 2008.

\bibitem{pytorchdeeplab}
Kazuto Nakashima.
\newblock {DeepLab with PyTorch}.
\newblock \url{https://github.com/kazuto1011/deeplab-pytorch}.

\bibitem{sawatzky2019harvesting}
Johann Sawatzky, Debayan Banerjee, and Juergen Gall.
\newblock Harvesting information from captions for weakly supervised semantic
  segmentation.
\newblock In {\em ICCV Workshop}, 2019.

\bibitem{selvaraju2017grad}
Ramprasaath~R Selvaraju, Michael Cogswell, Abhishek Das, Ramakrishna Vedantam,
  Devi Parikh, and Dhruv Batra.
\newblock Grad-cam: Visual explanations from deep networks via gradient-based
  localization.
\newblock In {\em ICCV}, 2017.

\bibitem{shen2018bootstrapping}
Tong Shen, Guosheng Lin, Chunhua Shen, and Ian Reid.
\newblock Bootstrapping the performance of webly supervised semantic
  segmentation.
\newblock In {\em CVPR}, 2018.

\bibitem{Shimoda_2019_ICCV}
Wataru Shimoda and Keiji Yanai.
\newblock Self-supervised difference detection for weakly-supervised semantic
  segmentation.
\newblock In {\em ICCV}, 2019.

\bibitem{song2019box}
Chunfeng Song, Yan Huang, Wanli Ouyang, and Liang Wang.
\newblock Box-driven class-wise region masking and filling rate guided loss for
  weakly supervised semantic segmentation.
\newblock In {\em CVPR}, 2019.

\bibitem{su2021context}
Yukun Su, Ruizhou Sun, Guosheng Lin, and Qingyao Wu.
\newblock Context decoupling augmentation for weakly supervised semantic
  segmentation.
\newblock {\em ICCV}, 2021.

\bibitem{sun2020mining}
Guolei Sun, Wenguan Wang, Jifeng Dai, and Luc Van~Gool.
\newblock Mining cross-image semantics for weakly supervised semantic
  segmentation.
\newblock In {\em ECCV}, 2020.

\bibitem{tang2018normalized}
Meng Tang, Abdelaziz Djelouah, Federico Perazzi, Yuri Boykov, and Christopher
  Schroers.
\newblock Normalized cut loss for weakly-supervised cnn segmentation.
\newblock In {\em CVPR}, 2018.

\bibitem{vilar2021extracting}
Daniel~R Vilar and Claudio~A Perez.
\newblock Extracting structured supervision from captions for weakly supervised
  semantic segmentation.
\newblock {\em IEEE Access}, 2021.

\bibitem{wang2017learning}
Lijun Wang, Huchuan Lu, Yifan Wang, Mengyang Feng, Dong Wang, Baocai Yin, and
  Xiang Ruan.
\newblock Learning to detect salient objects with image-level supervision.
\newblock In {\em CVPR}, 2017.

\bibitem{wang2018weakly}
Xiang Wang, Shaodi You, Xi Li, and Huimin Ma.
\newblock Weakly-supervised semantic segmentation by iteratively mining common
  object features.
\newblock In {\em CVPR}, 2018.

\bibitem{wang2020self}
Yude Wang, Jie Zhang, Meina Kan, Shiguang Shan, and Xilin Chen.
\newblock Self-supervised equivariant attention mechanism for weakly supervised
  semantic segmentation.
\newblock In {\em CVPR}, 2020.

\bibitem{wu2021embedded}
Tong Wu, Junshi Huang, Guangyu Gao, Xiaoming Wei, Xiaolin Wei, Xuan Luo, and
  Chi~Harold Liu.
\newblock Embedded discriminative attention mechanism for weakly supervised
  semantic segmentation.
\newblock In {\em CVPR}, 2021.

\bibitem{wu2019wider}
Zifeng Wu, Chunhua Shen, and Anton Van Den~Hengel.
\newblock Wider or deeper: Revisiting the resnet model for visual recognition.
\newblock {\em Pattern Recognition}, 2019.

\bibitem{xu2021leveraging}
Lian Xu, Wanli Ouyang, Mohammed Bennamoun, Farid Boussaid, Ferdous Sohel, and
  Dan Xu.
\newblock Leveraging auxiliary tasks with affinity learning for weakly
  supervised semantic segmentation.
\newblock In {\em ICCV}, 2021.

\bibitem{yao2021nonsalient}
Yazhou Yao, Tao Chen, Guosen Xie, Chuanyi Zhang, Fumin Shen, Qi Wu, Zhenmin
  Tang, and Jian Zhang.
\newblock Non-salient region object mining for weakly supervised semantic
  segmentation.
\newblock In {\em CVPR}, 2021.

\bibitem{zhang2021affinity}
Bingfeng Zhang, Jimin Xiao, Jianbo Jiao, Yunchao Wei, and Yao Zhao.
\newblock Affinity attention graph neural network for weakly supervised
  semantic segmentation.
\newblock {\em TPAMI}, 2021.

\bibitem{zhang2020causal}
Dong Zhang, Hanwang Zhang, Jinhui Tang, Xiansheng Hua, and Qianru Sun.
\newblock Causal intervention for weakly-supervised semantic segmentation.
\newblock In {\em NeurIPS}, 2020.

\bibitem{zhou2016learning}
Bolei Zhou, Aditya Khosla, Agata Lapedriza, Aude Oliva, and Antonio Torralba.
\newblock Learning deep features for discriminative localization.
\newblock In {\em CVPR}, 2016.

\end{thebibliography}
}
\setcounter{section}{0}
\renewcommand\thesection{\Alph{section}}
\setcounter{table}{0}
\renewcommand{\thetable}{A\arabic{table}}
\setcounter{figure}{0}
\renewcommand{\thefigure}{A\arabic{figure}}
\clearpage
\section{Appendix}
\subsection{Reproducibility}
Our implementations are based on deeplab-pytorch~\cite{pytorchdeeplab} for the ResNet-101 backbone and MMsegmentation~\cite{mmseg2020} for the Wide ResNet-38 backbone.

\subsection{Additional Analysis}
\begin{figure*}[t]
\centering
\includegraphics[width=\linewidth]{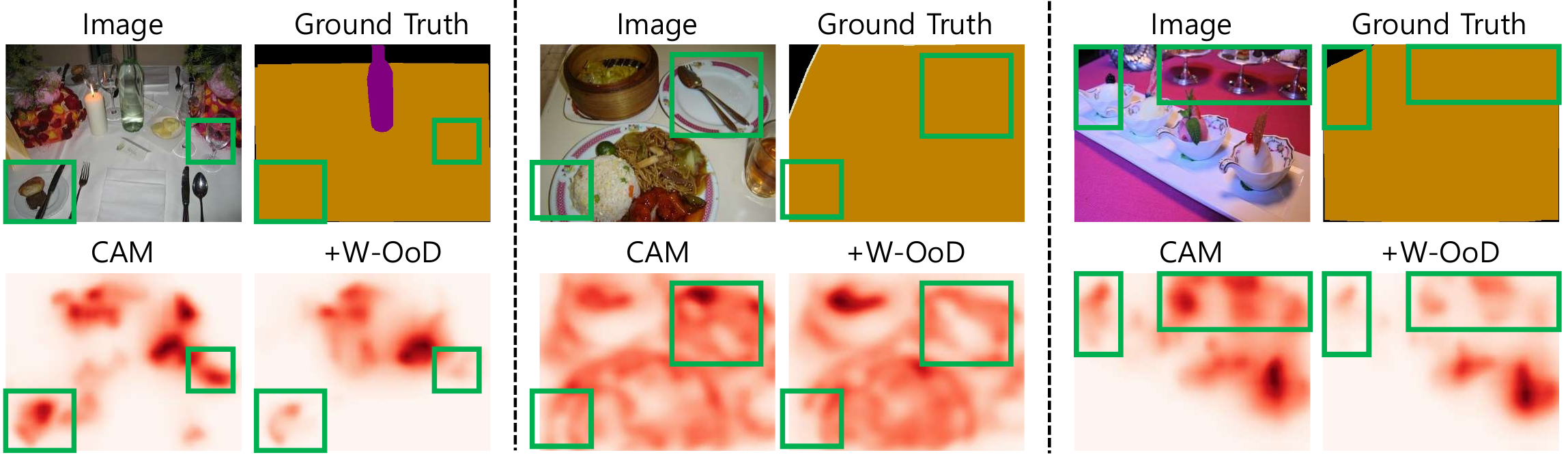}
\vspace{-2em}
\caption{\label{dining_mistake} Example where the objects on the dining table are not identified as foreground by our method.}
\vspace{-1.em}
\end{figure*}

\textbf{Examples of dining table:} We provide some examples of localization maps for ``dining table'' in Figure~\ref{dining_mistake},  mentioned in Section \textcolor{red}{4.3.3} in the main paper.

\begin{figure*}[t]
\centering
\includegraphics[width=\linewidth]{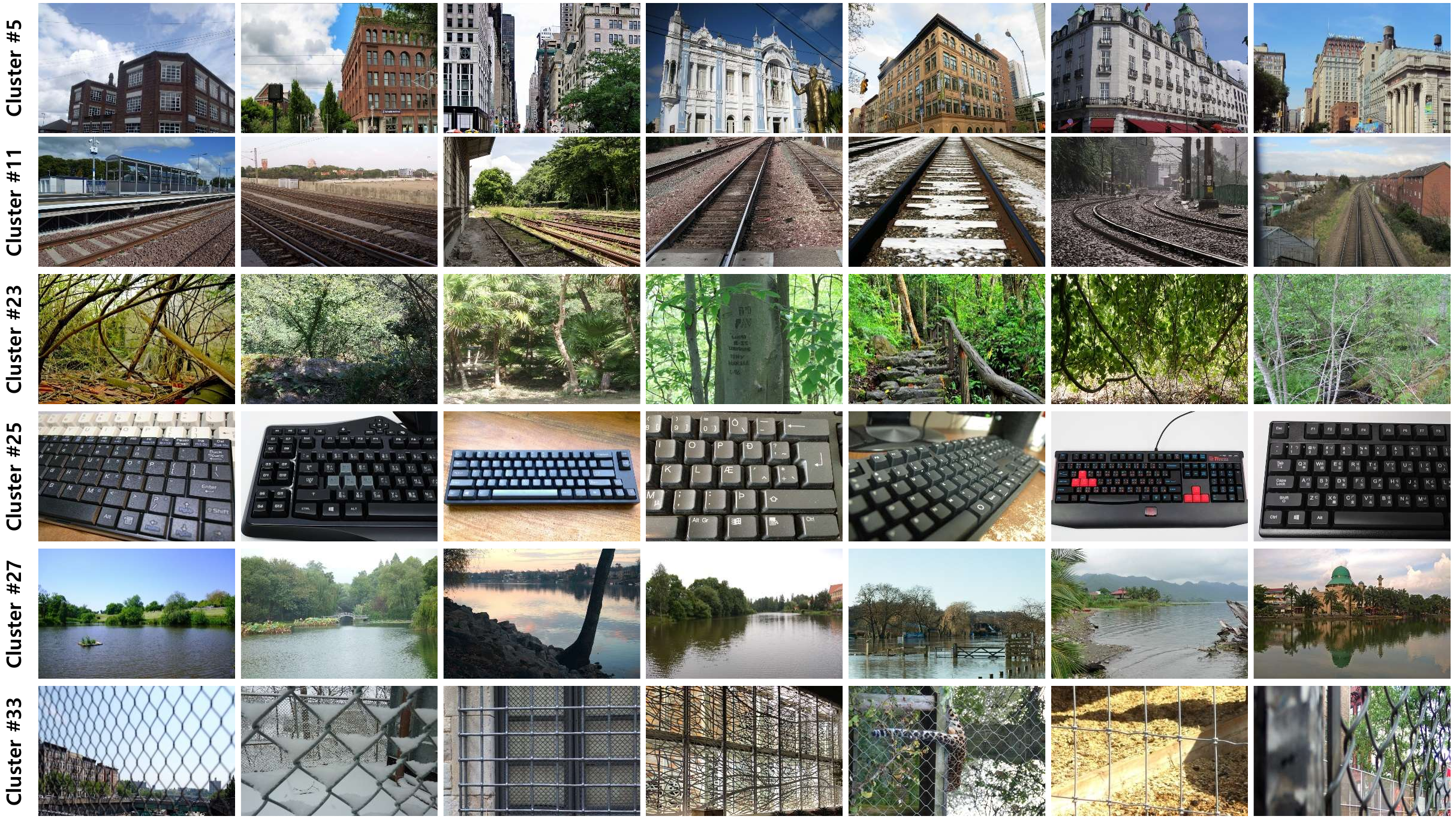}
\vspace{-2em}
\caption{\label{ood_samples} Examples of OoD samples for each cluster, obtained by the K-means clustering algorithm..}
\vspace{-1.em}
\end{figure*}

\textbf{Clustered OoD samples:} We provide examples of OoD samples clustered by the K-means clustering algorithm in Figure~\ref{ood_samples}.

\begin{figure*}[t]
\centering
\includegraphics[width=1\linewidth]{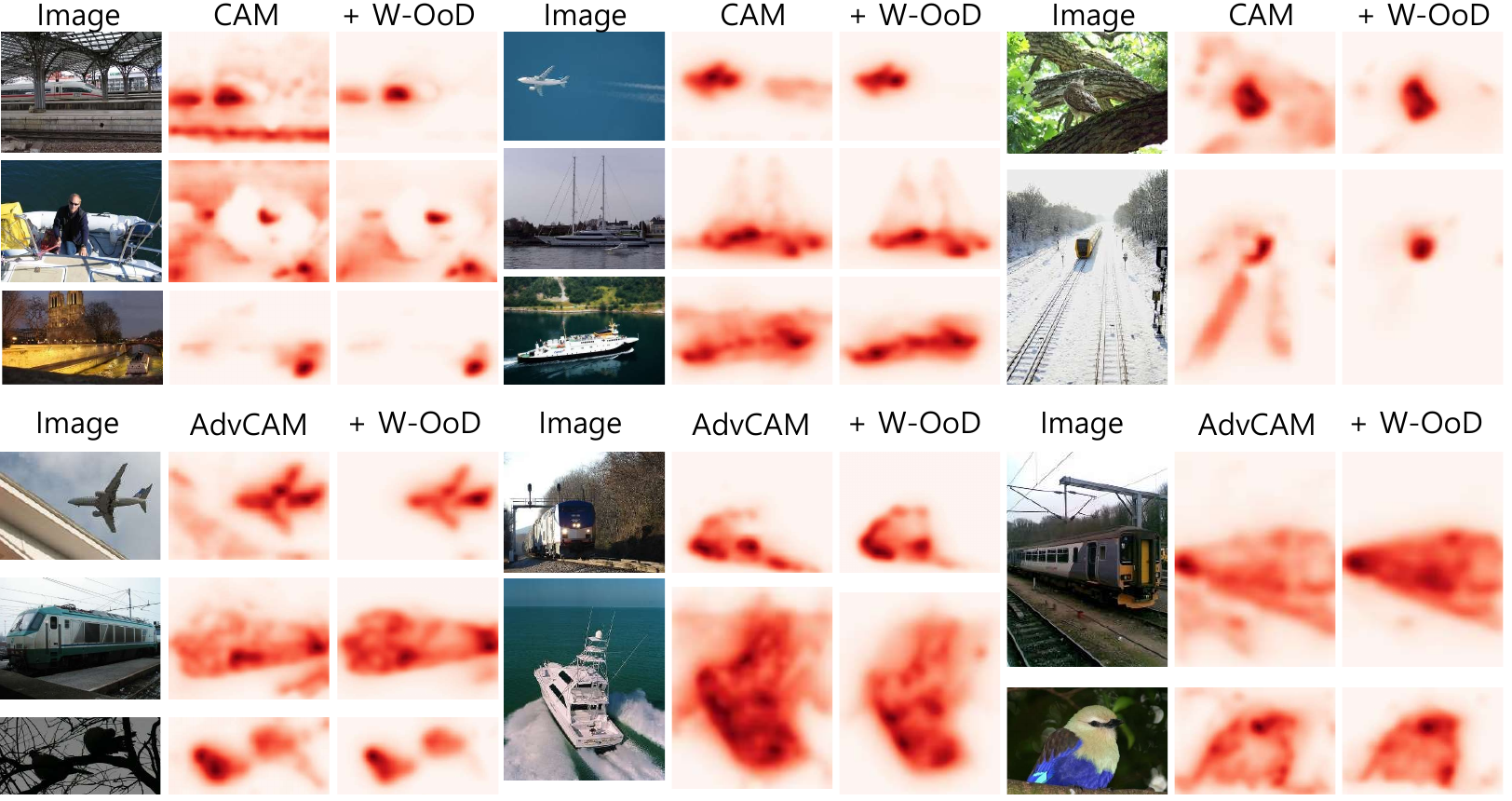}
\vspace{-1em}
\caption{\label{cam_samples_appendix} Examples of localization maps obtained from CAM and CAM+W-OoD (upper), and AdvCAM~\cite{lee2021anti} and AdvCAM+W-OoD (lower).
}
\vspace{-1em}
\end{figure*}

\begin{figure*}[t]
\centering
\includegraphics[width=1\linewidth]{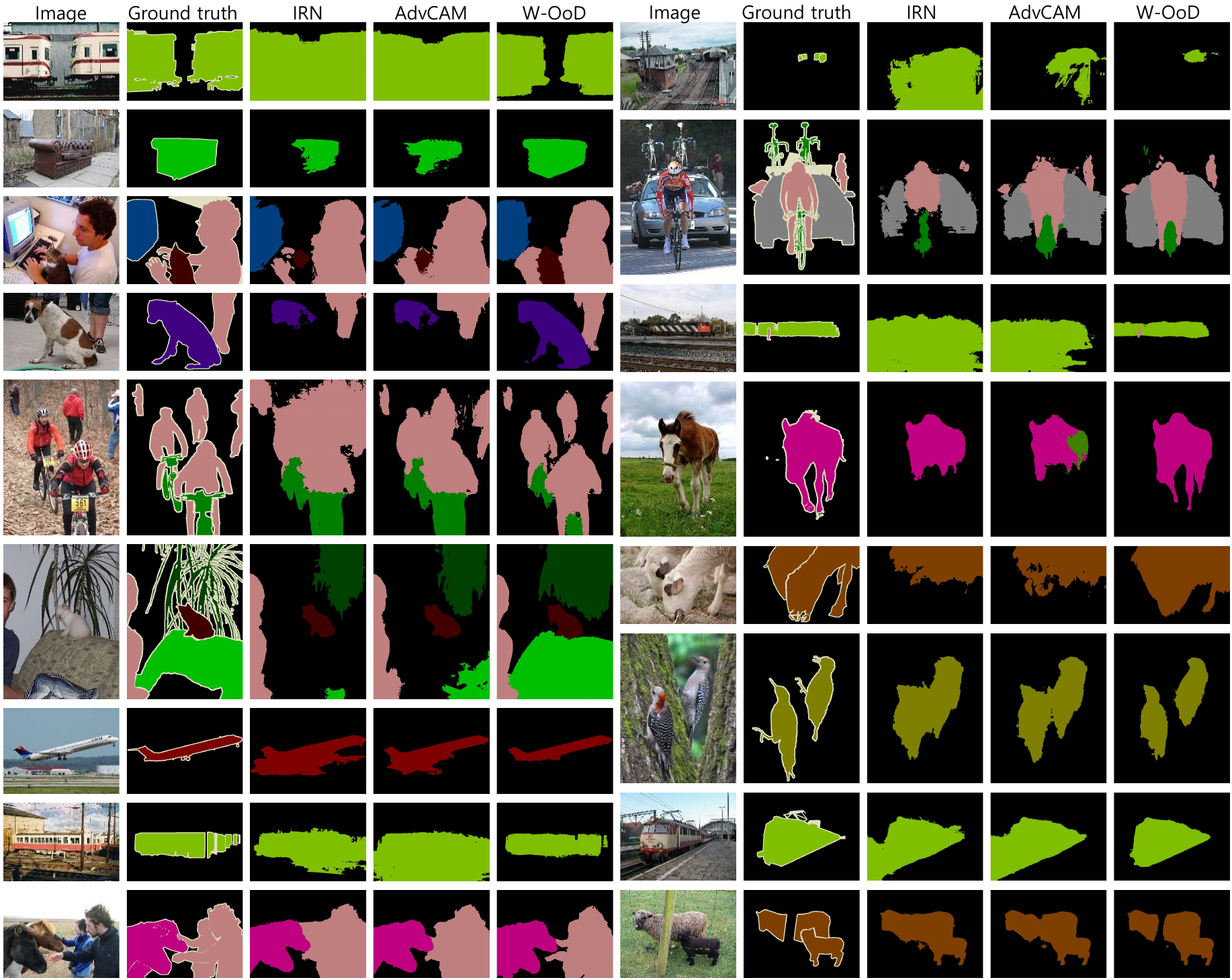}
\vspace{-1em}
\caption{\label{seg_samples_appendix} Examples of segmentation masks obtained by IRN~\cite{ahn2019weakly}, AdvCAM~\cite{lee2021anti}, and our method.
}
\vspace{-1em}
\end{figure*}

\textbf{More examples:} Figure~\ref{cam_samples_appendix} presents examples of localization maps obtained by IRN~\cite{ahn2019weakly} and our method, for the PASCAL VOC dataset.
Figure~\ref{seg_samples_appendix} shows examples of segmentation maps predicted by IRN~\cite{ahn2019weakly}, AdvCAM~\cite{lee2021anti}, and our method.

\textbf{Hyper-parameter analysis:} We analyze the sensitivity of the mIoU of the initial seed to $\tau$ and $\lambda$, hyper-parameters involved in the W-OoD training.
Since $\tau$ and $\lambda$ are dependent on each other, they must be searched jointly.
As the value of $\tau$ increases, $\mathcal{K}$ increases, so the value of $\mathcal{L}_\text{d}$ increases. Therefore, $\lambda$ must decrease accordingly.

\textbf{Segmentation results in mIoU with smaller $|\mathcal{D}_\text{ood}|$:}
In the main paper, we provide the quality of the initial seed by using smaller $|\mathcal{D}_\text{ood}|$. We here provide the final segmentation results with smaller $|\mathcal{D}_\text{ood}|$ in Table~\ref{table_smaller}.

\textbf{Comparison of per-class mIoU scores:}
Table~\ref{class-specific-results} shows the per-class mIoU of the final segmentation obtained by our method and recently produced methods.

\newpage
\begin{table}[htbp]
\caption{Segmentation results with smaller $|\mathcal{D}_\text{ood}|$ on Pascal VOC 2012 \textit{val} set.}\label{table_smaller}
  \centering
    \begin{tabular}{lcccc}
     \Xhline{1pt}
    { $|\mathcal{D}_\text{ood}|$}  &  0  & 20 & 500 & 5190\\ 
    \hline 
    mIoU & 67.5  &   68.1 & 69.0  & 69.8 \\
    \Xhline{1pt}
    \end{tabular}
\end{table}
\begin{table}[htbp]
  \centering
  \caption{Effect of the values of two hyper-parameters $\tau$ and $\lambda$.}
    \begin{tabular}{ccccc}
    \Xhline{1pt}
    $\tau$   & \multicolumn{4}{c}{$\lambda$} \\
\hline\hline \\[-0.9em]
\multicolumn{1}{c}{\multirow{2}[0]{*}{10}} & 0.01  & 0.015 & 0.02  & 0.025 \\
    \multicolumn{1}{c}{} & 52.5  & 52.6  & 53.1  & 52.2 \\
    \hline \\[-0.9em]
    \multicolumn{1}{c}{\multirow{2}[0]{*}{20}} &  0.003     &  0.005       &   0.007    &  0.01\\
    \multicolumn{1}{c}{} &   52.2    &  52.7   &      53.3 & 52.3 \\
        \hline \\[-0.9em]

    \multicolumn{1}{c}{\multirow{2}[0]{*}{30}} &    0.001   &  0.002     &  0.003     & 0.004 \\
    \multicolumn{1}{c}{} &    50.8   &   52.1    &   52.5    &  52.6\\
        \hline \\[-0.9em]

    \multicolumn{1}{c}{\multirow{2}[0]{*}{40}} &   0.0003    & 0.0005      &     0.0007  & 0.001 \\
    \multicolumn{1}{c}{} &    50.8   &   50.8    &    51.0   & 50.7 \\
    
    \Xhline{1pt}
    \end{tabular}%
  \label{table_tau_lambda}%
\end{table}%


\renewcommand{\tabcolsep}{2pt}

\begin{table*}[t]
  \caption{Comparison of per-class mIoU scores for the Pascal VOC 2012 dataset.}
  \centering
  \begin{adjustbox}{max width=\textwidth}
    \begin{tabular}{lccccccccccccccccccccc|c}
    
        \Xhline{1pt}

        \\[-0.95em]
     & bkg& aero  & bike  & bird  & boat  & bottle & bus   & car   & cat   & chair & cow   & table & dog   & horse & motor & person & plant & sheep & sofa  & train & tv  ~ & mIOU \\
   
    \hline
    \hline
    \\[-0.9em]
   \multicolumn{22}{l}{Results on PASCAL VOC 2012 validation images:}\\
    PSA~\cite{ahn2018learning}&   88.2    &   68.2    &   30.6    &   81.1    &   49.6   &   61.0    &    77.8  &    66.1   &    75.1   &  29.0     &   66.0    &   40.2    &    80.4 &    62.0   &   70.4    &  73.7 &  42.5   &70.7  & 42.6  &  68.1 &   51.6~ &  61.7\\
    CIAN~\cite{fan2018cian}&   88.2    &   79.5    &   32.6    &   75.7    &   56.8   &   72.1    &    85.3  &    72.9   &    81.7   &  27.6     &   73.3    &   39.8    &    76.4 &    77.0   &   74.9    &  66.8 &  46.6   &    81.0  & 29.1  &  60.4 &   53.3~ &  64.3\\
    SEAM~\cite{wang2020self} &   88.8    &   68.5    &   33.3    &  85.7    &   40.4   &   67.3    &    78.9  &    76.3   &    81.9   &  29.1     &   75.5    &   48.1    &    79.9 &    73.8   &   71.4    &  75.2 &  48.9   &    79.8  & 40.9  &  58.2 &   53.0~ &  64.5\\
    FickleNet~\cite{lee2019ficklenet} &   89.5    &   76.6    &   32.6    &   74.6    &   51.5   &   71.1    &    83.4  &    74.4   &    83.6   &  24.1     &   73.4    &   47.4    &    78.2 &    74.0   &   68.8    &  73.2 &  47.8   &    79.9  & 37.0  &  57.3 &   64.6~ &  64.9\\
    SSDD~\cite{Shimoda_2019_ICCV} &   89.0    &   62.5    &   28.9    &   83.7    &   52.9   &   59.5    &    77.6  &    73.7   &    87.0   &  34.0     &   83.7    &   47.6    &    84.1 &    77.0   &   73.9    &  69.6 &  29.8   &    84.0  & 43.2  &  68.0 &   53.4~ &  64.9\\
    BBAM~\cite{lee2021bbam} &   92.7    &   80.6    &   33.8    &  83.7    &   64.9   &   75.5    &    91.3  &    80.4   &    88.3   &  37.0     &   83.3    &   62.5    &    84.6 &    80.8   &   74.7    &  80.0 &  61.6   &    84.5  & 48.6  &  85.8 &   71.8~ &  73.7\\
    AdvCAM~\cite{lee2021anti} &   89.5    &   76.9    &   33.5    &   80.3    &   63.7   &   68.6    &    89.7  &   77.9   &    87.6   &  31.6     &   77.2    &   36.2    &    82.6 &    78.7   &   73.5    &  69.8 &  51.9   &    81.9  & 43.8  &  70.9 &   52.6  &  67.5\\
    W-OoD (ResNet-101) &   91.2    &   80.1    &   34.0    &   82.5    &   68.5   &   72.9    &    90.3  &   80.8   &    89.3   &  32.3     &   78.9    &   31.1    &    83.6 &    79.2   &   75.4    &  74.4 &  58.0   &    81.9  & 45.2  &  81.3 &   54.8  &  69.8\\
    W-OoD (WideResNet-38) &   91.0    &   80.1    &   34.1    &   88.1    &   64.8   &   68.3    &    87.4  &   84.4   &    89.8   &  30.1     &   87.8    &   34.7    &    87.5 &    85.9   &   79.8    &  75.0 &  56.4   &    84.5  & 47.8  &  80.4 &   46.4  &  70.7\\
    \hline\\[-0.9em]
    \multicolumn{22}{l}{Results on PASCAL VOC 2012 test images:}\\
    PSA~\cite{ahn2018learning}&   89.1    &   70.6    &   31.6    &   77.2    &   42.2   &   68.9    &    79.1  &    66.5   &    74.9   &  29.6     &   68.7    &   56.1    &    82.1 &    64.8   &   78.6    &  73.5 &  50.8   & 70.7  & 47.7  &  63.9 &   51.1~ &  63.7\\
    FickleNet~\cite{lee2019ficklenet} &   90.3    &   77.0    &   35.2    &   76.0    &   54.2   &   64.3    &    76.6  &    76.1   &    80.2   &  25.7     &   68.6    &   50.2    &    74.6&    71.8   &   78.3    &  69.5 &  53.8   &    76.5  & 41.8  &  70.0 &   54.2~ &  65.3\\
    SSDD~\cite{Shimoda_2019_ICCV} &   89.0    &   62.5    &   28.9    &   83.7    &   52.9   &   59.5    &    77.6  &    73.7   &    87.0   &  34.0     &   83.7    &   47.6    &    84.1 &    77.0   &   73.9    &  69.6 &  29.8   &    84.0  & 43.2  &  68.0 &   53.4~ &  64.9\\
    BBAM~\cite{lee2021bbam} &   92.8    &   83.5    &   33.4    &   88.9    &   61.8   &   72.8    &    90.3  &    83.5   &    87.6   &  34.7     &   82.9    &   66.1    &    83.9 &    81.1   &   78.3    &  77.4 &  55.2   &    86.7  & 58.5  &  81.5 &   66.4  &  73.7\\
    AdvCAM~\cite{lee2021anti} &   89.3    &   79.3    &   32.5    &   80.2    &   56.3   &   62.8    &    87.2  &   80.8   &    87.0   &  28.9     &   78.3    &   41.3    &    82.1 &    80.6  &   77.7    &  68.5 &  51.2   &    80.8  & 55.3  &  60.8 &   48.1  &  67.1$^1$\\
    W-OoD (ResNet-101) &   91.4    &   85.3    &   32.8    &   79.8    &   59.0   &   68.4    &    88.1  &   82.2   &    88.3   &  27.4     &   76.7    &   38.7    &    84.3 &    81.1   &   80.3    &  72.8 &  57.8   &    82.4  & 59.5  &  79.5 &   52.6  &  69.9$^2$\\
    W-OoD (WideResNet-38) &   90.9    &   83.1    &   35.6    &   89.0    &   61.5   &   63.0    &    86.2  &   80.8   &    89.9   &  29.6     &   79.6    &   40.1    &    82.1 &    81.0   &   82.6    &  74.0 &  60.1   &    85.3  & 58.0  &  71.9 &   47.0  &  70.1$^3$\\
        \Xhline{1pt}
    \end{tabular}%
  \end{adjustbox}%





  \label{class-specific-results}%
\end{table*}%

\end{document}